\documentclass[12pt]{article}
\usepackage{fullpage}

\usepackage{tikz}
\usetikzlibrary{arrows.meta,positioning,calc}

\usepackage[T1]{fontenc}
\usepackage[utf8]{inputenc}
\usepackage{lmodern}
\usepackage{microtype}
\usepackage{amsmath,amssymb}
\usepackage{graphicx}
\usepackage{booktabs}
\usepackage{endnotes}
\usepackage{hyperref}
\usepackage{url}
\usepackage{color}
\usepackage{xcolor}
\usepackage{ulem}
\usepackage{enumitem}
\usepackage{marvosym}
\usepackage{amsfonts}

\usepackage{xcolor}

\usepackage[full]{textcomp}
\usepackage[osf]{newtxtext}

\DeclareMathAlphabet{\catsymbfont}{U}{rsfs}{m}{n}

\newcounter{lwcount}

\newcounter{macount}

\newcounter{nscount}

\newcounter{rwcount}

\newcommand{\editorialdecision}[1]{{Editorial Decision:} #1}
\newcommand{\refereereportlinktext}[2]{submission-#1#2-reviews.pdf}
\newcommand{\refereereportlink}[3]{\href{https://github.com/1stproof/batch-2/blob/main/batch-2-reviews/problem-#1/submission-#3/submission-#2#3-reviews.pdf}{\refereereportlinktext{#2}{#3}}}

\title{First Proof Second Batch}
\date{June 10, 2026}
\author{}
\begin{document}
\maketitle
\bigskip

\begin{center}
{\small
\setlength{\tabcolsep}{3em}
\begin{tabular}{cc}
Mohammed Abouzaid\endnote{\texttt{abouzaid@stanford.edu}} &
Nikhil Srivastava\endnote{\texttt{nikhil@math.berkeley.edu}} \\[2pt]
\textit{Stanford University} &
\textit{University of California, Berkeley} \\[12pt]
Rachel Ward\endnote{\texttt{rward@math.utexas.edu}} &
Lauren Williams\endnote{\texttt{williams@math.harvard.edu}} \\[2pt]
\textit{University of Texas at Austin} &
\textit{Harvard University} \\
\end{tabular}
}
\end{center}

\bigskip

{\small\noindent Problem Contributors:}

\smallskip

{\small
\begin{enumerate}[leftmargin=2.4em,itemsep=4pt,label=\textbf{\arabic*.}]
\item Dariusz Kalociński, \textit{Polish Academy of Sciences}\endnote{\texttt{Dariusz.Kalocinski@ipipan.waw.pl}}; Theodore A.\ Slaman, \textit{UC Berkeley}\endnote{\texttt{slaman@math.berkeley.edu}}
\item Richard Schwartz, \textit{Brown University}\endnote{\texttt{richard\_schwartz@brown.edu}}
\item Aleksa Milojevi\'c, \textit{ETH Zürich}\endnote{\texttt{aleksa.milojevic@math.ethz.ch}}; Benny Sudakov, \textit{ETH Zürich}\endnote{\texttt{benjamin.sudakov@math.ethz.ch}}
\item Larry Guth, \textit{MIT}\endnote{\texttt{lguth@math.mit.edu}}
\item Oleg Butkovsky, \textit{Weierstrass Institute}\endnote{\texttt{oleg.butkovsky (at) wias (dash) berlin (dot) de}}; Jonathan Mattingly, \textit{Duke University}\endnote{\texttt{jonathan.mattingly@duke.edu}}; Lorenzo Zambotti, \textit{Sorbonne Universit\'e}\endnote{\texttt{zambotti@lpsm.paris}}
\item Joshua Evan Greene, \textit{Boston College}\endnote{\texttt{joshua.greene@bc.edu}}; Duncan McCoy, \textit{Université du Québec à Montréal}\endnote{\texttt{mc\_coy.duncan@uqam.ca}}
\item Sucharit Sarkar, \textit{UCLA}\endnote{\texttt{sucharit@math.ucla.edu}}
\item Sam Payne, \textit{University of Michigan}\endnote{\texttt{sdpayne@umich.edu}}; Jidong (Jayden) Wang, \textit{University of Michigan}\endnote{\texttt{jidongw@umich.edu}}
\item Sylvie Corteel, \textit{UC Berkeley}\endnote{\texttt{corteel@berkeley.edu}}; John Lentfer, \textit{UC Berkeley}\endnote{\texttt{jlentfer@berkeley.edu}}
\item Srivatsav Kunnawalkam Elayavalli, \textit{University of Maryland}\endnote{\texttt{sriva@umd.edu}}
\end{enumerate}
}

\begin{abstract}
    To assess the ability of current AI systems to correctly solve research-level mathematics problems, we tested several AI systems on a set of ten problems in a broad range of mathematical fields; these problems arose naturally in the research process of the contributors.  This document includes the problems, our methodology, and the results of our testing.  We provide links to supplementary documents including the human solutions, the AI-generated solutions, and the referee reports and logs for the AI-generated solutions.
    \end{abstract}

\newpage

\tableofcontents

\section{Introduction}

The goal of our ongoing First Proof project 
is to provide accurate information to mathematicians and to the public about the capabilities of AI systems in the context of mathematical research. 
The first batch of First Proof \cite{FirstProof} was an informal collaborative experiment, consisting of 10 problems that could be used to measure the ability of AI systems to provide proofs of mathematical statements.  More specifically, these problems were statements which had come up naturally in the mathematical research of the authors -- but whose proofs had not appeared on the internet or been published.  In the first batch, we made our problems public, allowed the community to work on them for a week, and then released the solutions. We did not enforce rules for how the community should engage with the problems, and we did not grade the solutions.  Instead, we used this first batch as an experiment which informed our design of the second batch of problems -- a batch that we have intended as a more formal \textbf{benchmark}.

This document describes how we created, tested,
and graded the second batch of problems from March to June 2026. Our process was overseen by the {First Proof Foundation}.  We invited commercial companies and academic teams to participate in our benchmark, and, guided by our goal of transparency (\textit{cf.} Section \ref{sec:methods}) we ultimately accepted to test and grade {\bf four systems}: OpenAI's ChatGPT 5.5 Pro, as well as \textbf{harnesses} (code which calls publicly available models via API) created by academic teams from ETH Zurich/Aarhus University, UCLA, and Princeton.  In the coming weeks, we will also release a separate smaller collection of problems to allow for community experimentation.

\subsection{What is mathematics?}
Before discussing more details, it is worth taking a moment to discuss
what mathematics is and what mathematicians do.  Mathematics is an inherently creative discipline, in which mathematicians push the boundary of mathematics by asking new questions and developing frameworks to answer these questions.  Henri Poincaré described mathematical creation as a largely unconscious imaginative process, guided by aesthetic selection: {\it It is by logic that we prove, but by intuition that we discover.}  Indeed, the work of discovering novel and interesting conjectures, as well as the right conceptual definitions and frameworks for approaching such conjectures, is at least as important as the work of proving such conjectures.  One can make an analogy with art: the first step is to conceive of the idea; only having done so can one execute the work itself. 
As Sol LeWitt wrote: {\it in conceptual art the idea or concept is the most important aspect of the work \dots the idea becomes a machine that makes the art.}  
\begin{figure}[h]
\centering
\begin{tikzpicture}[
    box/.style={
        draw,
        rectangle,
        rounded corners,
        minimum width=3.2cm,
        minimum height=1cm,
        align=center,
        inner sep=3pt
    },
    arrow/.style={->, thick, >=Latex},
    node distance=1.2cm
]

\node[box] (q) {asking new\\questions};
\node[box, right=of q] (f) {developing frameworks\\to answer them};
\node[box, right=of f] (s) {solving questions\\using frameworks};

\draw[arrow] (q) -- (f);
\draw[arrow] (f) -- (s);

\draw[arrow]
    (s.south) .. controls +(0,-1.6) and +(0,-1.6) .. (q.south);

\end{tikzpicture}
\caption{The mathematical research cycle.}
\label{fig:math-cycle}
\end{figure}
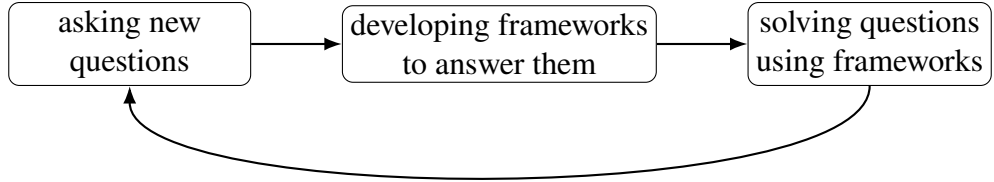

At present we do not have concrete proposals for devising experiments that would measure the ability of AI systems to produce interesting new conjectures, or useful mathematical definitions and frameworks.  In this second batch, we have thus, as in the first batch, restricted our attention to assessing mathematical proofs that AI systems have created.  As in our initial experiment, we collected solved but unpublished mathematical problems from human mathematicians, which we posed as a challenge to AI systems.  

\subsection{Summary of Methodology}\label{sec:methods}
In keeping with the values of the academic math community, our guiding principle in designing the second batch was \textbf{transparency}.  We decided that in order to conduct an experiment in which we could have total clarity about prompts and human interaction, we needed to run all of the tests ourselves.  Since our goal is to inform mathematicians and the public about models that are in principle available to them\footnote{Given sufficient funding.} we decided to test only those AI systems that were public at the time of testing, or to test harnesses that call only publicly available models and would be released to the public.  We also committed to 
 releasing a full account of our process: the ten problems, the human-generated solutions, the AI-generated solutions, the cost of creating those solutions, our evaluations of those solutions,  the source code of the AI systems modulo publicly available models, and the logs produced by the AI systems.

This batch consists of a formal benchmark, whose results we report here, and an upcoming community experiment.
Our methodology for selecting, testing, and grading the $10$ problems comprising our benchmark is described in detail in Section \ref{sec:meth}. A high-level summary indicating differences between the first batch and the second batch benchmark is provided in Table \ref{tab:batch-comparison}. The most important differences are that in this second batch, we {\bf executed all the testing ourselves}, and -- thanks to the assistance of $30$ expert mathematicians -- we {\bf formally assessed the solutions} that the AI systems produced.

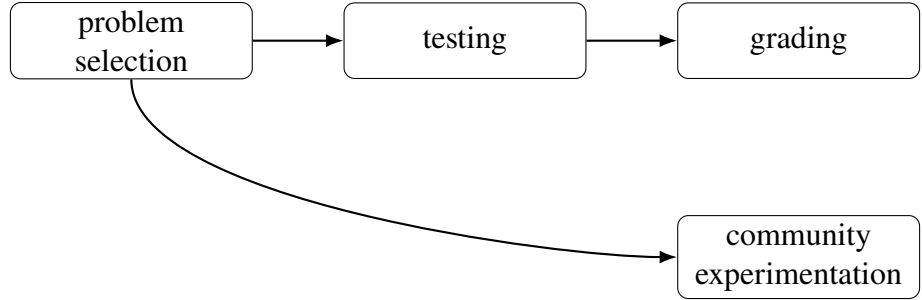
\begin{figure}[h]
\centering
\begin{tikzpicture}[
    box/.style={
        draw,
        rectangle,
        rounded corners,
        minimum width=3.2cm,
        minimum height=1cm,
        align=center,
        inner sep=3pt
    },
    arrow/.style={->, thick, >=Latex},
    node distance=1.2cm
]

\node[box] (p) {problem\\selection};
\node[box, right=of p] (t) {testing};
\node[box, right=of t] (g) {grading};

\node[box, below=1.8cm of g] (c) {community\\experimentation};

\draw[arrow] (p) -- (t);
\draw[arrow] (t) -- (g);

\draw[arrow]
(p.south) .. controls +(0,-1.6) and +(-1.6,0) .. (c.west);

\end{tikzpicture}
\caption{Overview of the First Proof Second Batch  methodology.}
\label{fig:methodology}
\end{figure}

\begin{table}[h]\label{table:meth}
\centering\small
\begin{tabular}{p{0.28\textwidth}p{0.32\textwidth}p{0.32\textwidth}}
\toprule
& \textbf{Batch 1 (February 2026)} & \textbf{Batch 2 Benchmark (March-June, 2026)} \\
\midrule
\multicolumn{3}{l}{\textbf{Problem Selection}} \\
\quad Number of problems & 10 & 10 \\
\quad Problem source & Research mathematicians & Research mathematicians \\
\quad Max proof length & 5 pages & 8 pages \\
\quad Preliminary AI testing & ChatGPT 5.2-Pro,  Gemini 3.0 Deep Think & ChatGPT 5.5-high/xhigh, Gemini 3.1 Pro, Opus 4.7 (via OpenRouter) \\
\midrule
\multicolumn{3}{l}{\textbf{Testing}} \\
\quad Who may participate & Open to anyone & Strict eligibility requirements (open-source harness
calling public models via API) \\
\quad Who runs the tests & Participants & First Proof \\
\quad Autonomy of AI systems & Not verified & Near-guaranteed \\
\midrule
\multicolumn{3}{l}{\textbf{Grading}} \\
\quad Process & Informal Zulip discussion & Double blind journal review model with $\sim$30 expert referees \\
\quad Correctness of solutions & Informal community consensus &  Each submission verified by 2-3 independent referees\\
\bottomrule
\end{tabular}
\caption{Similarities and differences between Batch 1 and Batch 2 Benchmark.
\label{tab:batch-comparison}}
\end{table}

\subsection{Summary and Discussion of Results}

We tested four systems on the Second Batch of ten problems, and had thirty-nine\footnote{System A did not produce a solution on Problem $6$ due to a technical issue; see Section \ref{sec:P6}} solutions that needed to be graded.  Each solution  was graded by
at least two referees who are experts in the mathematical area of the problem.  Using their  feedback, we rated each solution as either essentially flawless, requiring minor revisions, requiring major revisions, or rejected.  

Across all four systems, a combined total of \textbf{7 problems} received at least one passing grade (rated as essentially flawless or requiring only minor revisions).  Some solutions were rated as essentially publishable; notably, Problem 5 (stochastic PDE) was solved correctly by one system using a novel approach that differed from the human solution and impressed the referees. At the other extreme were problems which saw complete failure, most notably Problem 4 (metric geometry), where no system made substantial progress. There were an additional \textbf{2 problems} for which at least one submission was rated as requiring major revisions — meaning referees judged that the approach was potentially viable but that substantial human effort would be needed to repair it. Detailed results are presented in Section \ref{sec:results}, with summaries of referee recommendations appearing in Table \ref{tab:decisiongrid} and summaries of costs in Table \ref{tab:costs}.

Here are some additional observations about the AI systems and the solutions they generated:
\begin{itemize}
\item AI systems were sometimes able to produce novel arguments which were different from the human solutions, as in Problems 3, 5, and 9.
 \item AI systems tended to perform best when a problem was structurally similar to results already in the literature. 
 Problem 2  was analogous to a problem the author had previously solved and published; three AI systems solved Problem 2, all of them by translating the proof of the previously solved problem into the appropriate category.
\item A recurring theme in the referee reports was that AI solutions tended to handle routine parts of an argument in meticulous detail while glossing over the most difficult steps, sometimes asserting that a key claim follows from "standard arguments" without justification, or citing papers that do not actually contain the claimed results. 
 \item Despite the fact that the harnesses attempted to check citations, we saw a lot of failures on this front.  
 For instance, several of the solutions to Problem 2 borrowed phrasing from the author's previous paper even line by line, reusing its terminology (“T -patterns,” “bends”) and even its labels (B, T, D, H), but without citing the paper anywhere. If a human had submitted such a solution it would have been flagged for plagiarism.
\item It is possible for academic teams to design AI harnesses which can improve on the mathematical quality of the output of their base model(s). However, this improvement  comes at a substantial financial cost.
\end{itemize}

\subsection{What comes next} 

\paragraph{Community Experimentation (June and July, 2026)}

On June 17, 2026 we will launch an informal 
community experiment to allow the general public to test AI systems of their choice on a smaller, separate batch of problems. We will release two problems on June 17 and release their solutions one week later. 
In subsequent weeks we will release several more problems.
Each time we release solutions we will direct participants to a public Zulip where they can discuss the problems, their AI systems, and their solutions.  We will not formally verify autonomy or correctness, but request that participants include the following with their solutions (which will make it easier to interpret the results during the discussion):
\begin{itemize}
    \item  a complete transcript of the interaction with AI models, fully specifying the details of human involvement, and  
    \item a record of the resources used, such as number or cost of tokens.
\end{itemize}

\paragraph{Third Batch (August-October 2026)}
From August to October 2026 we will create, test, and grade a third batch of problems.  We expect the process to be  similar to the process we used for the second batch. 
We will provide more details about the third batch on \textbf{July 20}. 
\bigskip

\noindent{\bf Acknowledgements:~}
We are tremendously grateful to the mathematicians who  created the problems of Batch 2, and to the mathematicians who assessed the solutions.  This project would have been impossible without their expertise.

We would like to thank 
the Simons Institute for the Theory of Computing, which hosted our meeting in April to create the problems for Batch 2, with support from the Director's Opportunity Fund, as well as Harvard's Center for Mathematical Sciences and Applications (CMSA), which hosted the meeting in June to grade the AI-generated solutions.  We are also grateful to the Institute for Computer-Aided Reasoning in Mathematics (ICARM), which will host the Zulip for our community experiment.  In addition, this work has been supported by grants from the Survival and Flourishing Fund, and the AI For Math Fund, which is run by Renaissance Philanthropy with support from XTX Markets. 

First Proof has obtained unrestricted donations from Anthropic and from OpenAI, with pending funding from Google.org; see Section \ref{sec:funding} for information about how donations from AI companies are used.

\paragraph{Organization.} The remainder of this document contains: a description of our process for testing and grading; the ten problems; the four teams who entered Batch 2; their compute costs for working on the Batch 2 problems; and the results of our grading process.  In addition, our \href{https://1stproof.org}{website} hosts additional documents: the human-generated solutions, the AI-generated solutions,  the referee reports on those solutions, and the logs produced by the AI systems.

\section{Methodology}\label{sec:meth}
\subsection{Problem Selection (March to May, 2026)}
\label{sec:problemselection}
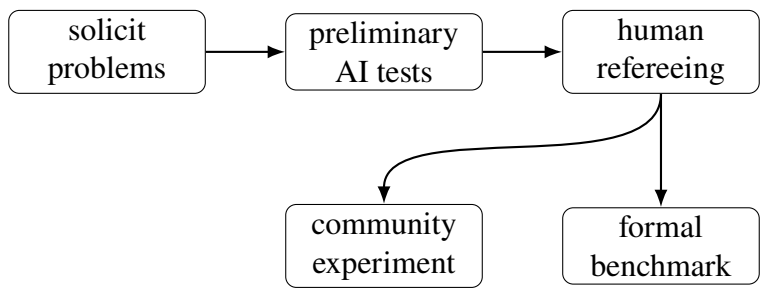
\begin{figure}[h]
\centering
\begin{tikzpicture}[
    box/.style={
        draw,
        rectangle,
        rounded corners,
        minimum width=2.6cm,
        minimum height=1cm,
        align=center,
        inner sep=3pt
    },
    arrow/.style={->, thick, >=Latex},
    node distance=1.05cm
]

\node[box] (s) {solicit\\problems};
\node[box, right=of s] (p) {preliminary\\AI tests};
\node[box, right=of p] (r) {human\\refereeing};
\node[box, below=1.5cm of r] (b) {formal\\benchmark};
\node[box, left=of b] (c) {community\\experiment};

\draw[arrow] (s) -- (p);
\draw[arrow] (p) -- (r);
\draw[arrow] (r) -- (b);
\draw[arrow]
    (r.south) .. controls +(0,-1.2) and +(0,1.2) .. (c.north);

\end{tikzpicture}
\caption{Problem selection process.}
\label{fig:problem-selection}
\end{figure}
\paragraph{Solicitation.} We solicited problems from mathematicians representing a wide range of mathematical fields as well as geographic locations (mostly in the United States for logistical reasons). We aimed to obtain mathematical problems which
involve a nonstandard insight to answer,
and which have a proof known to the mathematician of {\bf at most 8 pages} which had not appeared on the internet or in any public forum.

We asked that the authors:
\begin{itemize}[itemsep=1pt]
\item {not discuss} the candidate problems with any AI system before the problems were announced, except in a controlled zero data retention environment (see below).
\item produce a {\bf solution} of at most 8 pages which consists of a complete proof, with accurate references to the mathematical literature;
\item pre-register a {\bf descriptive statement}, written for a general audience of mathematicians, describing the (i) novelty of the argument, (ii) the general strategy of the proof, and (iii) an estimate of the difficulty of the problem, e.g. via the time that it has taken to solve the problem.
\item confirm that they will not undertake any employment or consulting work for AI companies while contributing to this project, and commit to not doing so during March-September, 2026. 
\end{itemize}

\paragraph{Preliminary AI Tests.} For each candidate problem, we wanted to have high confidence about the following: (i) the problem is comprehensible to an LLM. (ii) it cannot be solved by a standard approach (iii) it cannot be answered by appealing to a nearby proof in the literature.
To do so, we ran preliminary tests during April 23-30, 2026, in which the authors of the problems queried the following models with Zero Data Retention via Openrouter, with a 30 minute timeout: ChatGPT 5.4/5.5-high and -xhigh, Gemini 3.1 Pro, and Opus 4.7. These models {\bf did not solve any of the ten problems in this benchmark}.  Notably, we did not query ChatGPT 5.5 Pro as it was not available with Zero Data Retention.  
{\bf Two problems were discarded} at this stage: 
 one was discarded because the response was an immediate reduction to a result in the literature;
 another was eliminated at the request of the author, because an LLM-search identified a reference from which the result could be quickly derived.

\paragraph{Human Solution Refereeing.} We had all (human-generated) solutions go through a first round of refereeing by a single human expert referee --- suggested by the author of the problem --- to check clarity and correctness. Feedback from the referee reports was used to revise some of the solutions.

\paragraph{Selection.} We selected {\bf 10 problems for the formal benchmark}, and {\bf a smaller and separate set of problems for the informal community experiment} (which will begin on June 17, 2026) based on the following criteria: (i) balance across fields of mathematics (ii) balance across geographic locations (iii) balance across perceived difficulty and novelty by the authors.

\subsection{Testing  (May 28 - June 1, 2026)}
\begin{figure}[h]
\centering
\begin{tikzpicture}[
    box/.style={
        draw,
        rectangle,
        rounded corners,
        minimum width=2.6cm,
        minimum height=1cm,
        align=center,
        inner sep=3pt
    },
    arrow/.style={->, thick, >=Latex},
    node distance=1.05cm
]

\node[box] (e) {eligibility\\criteria};
\node[box, right=of e] (sel) {select\\systems};
\node[box, right=of sel] (v) {validate\\submitted code};
\node[box, below=1.5cm of v] (run) {run on First Proof AWS account};
\node[box, left=of run] (out) {retrieve outputs\\and logs};

\draw[arrow] (e) -- (sel);
\draw[arrow] (sel) -- (v);
\draw[arrow] (v) -- (run);
\draw[arrow] (run) -- (out);

\end{tikzpicture}
\caption{Testing process.}
\label{fig:testing}
\end{figure}
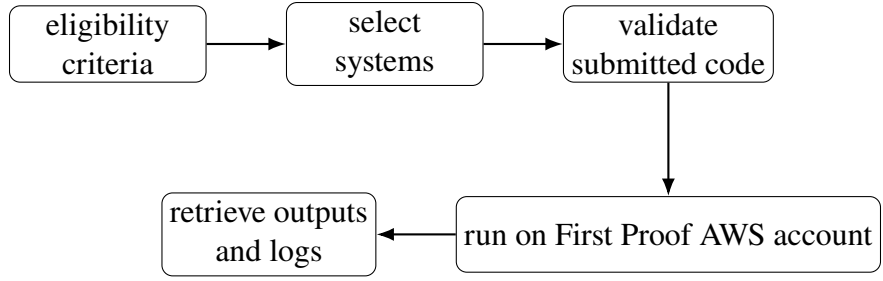
\paragraph{Eligibility.} We announced the second batch on March 14, 2026, and asked interested parties to contact us by April 14, 2026. Guided by our goal of transparency, we put the following constraints on AI systems to be tested in the second batch benchmark:
\begin{enumerate}
    \item The system must consist of code (which we will sometimes refer to as as a {\bf harness}) which calls publicly available models via API. Here, \textit{publicly available} means that any member of the public must be able to access the model as of May 28, 2026.
    \item The system must take as input ten math problems given as .tex source, solve them in one shot (i.e., with no further interaction), and output the results in within 24 hours.
    \item The system must log input, output, and reasoning tokens. We did not put an upper bound on the number of tokens allowed.
    \item First Proof must be allowed to publish all of the output, logs, and code on June 10, 2026.
    \item  The system must have performance that is comparable to the one-shot performance of the best publicly available models as of May, 2026. 
\end{enumerate}
Note that the above constraints allow as a special case a harness which just calls a public model using a one-shot prompt for each problem.

\paragraph{Selection of Systems to Test.} Based on the above criteria and discussions with interested parties, it was decided by May 28, 2026 that we would test four AI systems: ChatGPT 5.5 Pro from OpenAI, and three harnesses from academic teams at IMProofBench, UCLA, and Princeton. These are described in detail in Section \ref{sec:systems}.

\paragraph{Testing Implementation.} We required that each participating system be implemented as a complete, reproducible pipeline that could be deployed on standard cloud infrastructure, as outlined below. 

All systems processed a common JSON input file containing the 10 benchmark problems. First Proof owned the infrastructure: an AWS account, the input data, and a run.sh script that managed each run's full lifecycle. For each system, First Proof launched a fresh ephemeral CPU-only EC2 instance with 8 vCPUs, 64 GB of RAM, and 100 GB of disk storage. First Proof copied over the code and input file, built a Docker container, and executed it. Containers had outbound internet access for API calls and limited fetching like paper PDFs; inbound network access was restricted via security group to the operator's IP. The container read the input file and wrote its results to a designated output directory. Submitters provided any required API keys through a secure out-of-band channel. After the container exited or hit a 24-hour timeout, First Proof retrieved the outputs and logs, recorded run metadata, and terminated the instance.  The output was required to be ten solutions formatted as separate, properly compilable LaTeX documents, each up to at most 12 pages using  standard latex formats.   Successful testing required that submissions compile cleanly on Overleaf without modification. We made a good faith effort to resolve minor LaTeX compilation errors that arose during evaluation.

\subsection{Grading (June 4-8, 2026)}
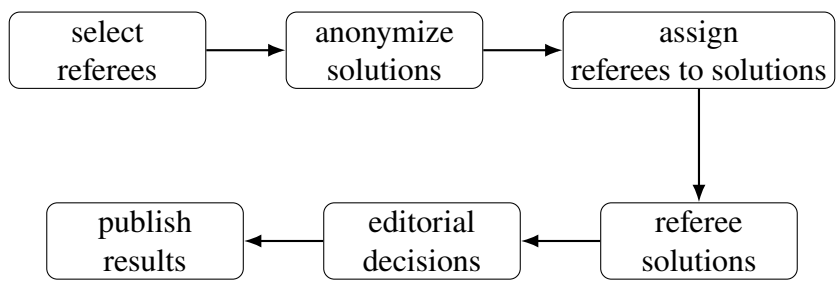
\begin{figure}[h]
\centering
\begin{tikzpicture}[
    box/.style={
        draw,
        rectangle,
        rounded corners,
        minimum width=2.6cm,
        minimum height=1cm,
        align=center,
        inner sep=3pt
    },
    arrow/.style={->, thick, >=Latex},
    node distance=1.05cm
]

\node[box] (ref) {select\\referees};
\node[box, right=of ref] (anon) {anonymize\\solutions};
\node[box, right=of anon] (assign) {assign\\referees to solutions};
\node[box, below=1.5cm of assign] (grade) {referee\\solutions};
\node[box, left=of grade] (ed) {editorial\\decisions};
\node[box, left=of ed] (pub) {publish\\results};

\draw[arrow] (ref) -- (anon);
\draw[arrow] (anon) -- (assign);
\draw[arrow] (assign) -- (grade);
\draw[arrow] (grade) -- (ed);
\draw[arrow] (ed) -- (pub);

\end{tikzpicture}
\caption{Grading process.}
\label{fig:grading}
\end{figure}
\paragraph{Selection of Referees.} We asked the authors of the problems to suggest a list of referees for each problem, that is, mathematicians who had expertise in the field and were qualified to grade the solutions. After several rounds of invitations, we were able to confirm 3 referees for each problem. The total number of confirmed referees across all problems was 30. 

\paragraph{Anonymization and Referee Assignment} The solutions produced by the AI systems were given randomized animal names (Ocelot, Badger, Marmot, etc.), with the identities of the corresponding AI systems visible only to the editors. Each submission was assigned $2$ or $3$ referees based on its perceived intricacy by the editors, which in turn was partially based on quick opinions from the problem authors.
 
\paragraph{Refereeing Process.} 
The anonymized solutions produced by the AI systems were graded by the human referees in a manner similar to the review process in mathematics journals, during an in-person meeting at Harvard's CMSA during June 4-5, 2026.

We asked the referees to give greatest weight to the \textbf{correctness of the solution}.
  
We also asked the referees to assess the \textbf{novelty of the solution}, i.e. whether it contained new ideas versus standard techniques, and to comment on the \textbf{quality of the exposition}, including the usage of \textbf{proper attributions} and \textbf{citations}.

We obtained at least two reports for each submission.  
Each solution was rated as:
\begin{itemize}[itemsep=1pt]
    \item essentially flawless (all disagreements are stylistic);
\item requires minor revisions;  (i.e. a few errors which can be readily corrected); 
\item requires major revisions;  (i.e. there is a flaw in the implementation of the strategy which requires significant work to address);
\item should be rejected (i.e. follows a strategy which is not clearly salvageable).
\end{itemize}

Referees were allowed to ask each other mathematical questions for clarification only and were instructed to make their own judgments. 

Some referees asked the editors for clarification about the boundaries between the four categories above. When in doubt about a submission {\em containing a mathematical error or gap}, we asked the reviewers to 
distinguish between minor and major revisions according to the following operational criterion: would the reviewer be so confident that a mathematician with the relevant domain knowledge could implement an appropriate fix that there would be {\em no need to review their correction}?  Based on the answer, the reviewer should assign a request for minor or major revisions. We conveyed that a comprehensible and mathematically correct solution should not be placed in a category lower than minor revisions, however bad the exposition or citations.

\paragraph{Editorial Decisions.} The Editorial Board read the referee reports and used the referee recommendations to produce a single verdict for each submission. 

\paragraph{Publication.} All problems, author solutions, author pre-registered statements, referee reports and editorial decisions, along with editorial reasoning, are published on June 10, 2026.

\subsection{Funding}\label{sec:funding}
We solicited unrestricted donations from commercial companies whose systems we considered testing. These donations are being allocated towards the costs required to create the problems and assess the solutions. For instance, we are providing modest research funding to each referee in recognization of their contribution.  On the other hand, these donations are {\it not} being used to compensate members of the editorial board or the board of directors.   

We also solicited additional funding from foundations in order to test systems from non-commercial entities.

The First Proof Foundation is registered as a 501(c)(3) corporation and it releases full reports of our expenses.

\newpage

\section{Problems}

\newcommand{\novsep}{\par\smallskip\noindent\rule{\linewidth}{0.4pt}\par\smallskip}
\newenvironment{novelty}[1]{%
  \novsep\noindent{\small\textbf{Pre-registered Statement} (\textit{#1}).}\par\nopagebreak\smallskip
  \small}{\par\novsep\normalsize}

\providecommand{\Vol}{\operatorname{Vol}}
\providecommand{\Dil}{\operatorname{Dil}}
\providecommand{\Trop}{\operatorname{Trop}}
\providecommand{\Dr}{\operatorname{Dr}}

\subsection{Problem 1: Ted Slaman and Dariusz Kalociński (computability theory)}

We consider infinite countable first-order structures, their automorphism groups, and the descriptive aspects of their representations. A representation $\mathcal{A}$ of such a structure is an instance of its isomorphism type in which the domain is the set of natural numbers $\mathbb{N}$, and the constants, functions, and relations are elements of $\mathbb{N}$, multi-variable functions from $\mathbb{N}$ to $\mathbb{N}$, and multi-place relations on $\mathbb{N}$, respectively. An automorphism of $\mathcal{A}$, or an isomorphism from $\mathcal{A}$ to $\mathcal{B}$, is a bijection of $\mathbb{N}$ preserving the evaluation of all constants, functions, and relations. We let $\mathit{AUT}(\mathcal{A})$ denote the automorphism group of $\mathcal{A}$.

\noindent\textbf{Definition.}
\begin{itemize}[itemsep=1pt]
\item $\mathcal{A}$ is \emph{computably $\mathit{AUT}$-countable} if every automorphism of $\mathcal{A}$ is computable relative to $\mathcal{A}$.
\item $\mathcal{A}$ is \emph{computably $\mathit{AUT}$-countable on a cone} if there is a $C\subseteq\mathbb{N}$ such that whenever $\mathcal{B}$ is isomorphic to $\mathcal{A}$, every automorphism of $\mathcal{B}$ is computable relative to $\mathcal{B}\oplus C$.
\end{itemize}

\noindent\textbf{Definition.} A function $f\colon\mathcal{A}\to\mathcal{A}$ is \emph{$\Sigma^{in}_1$-definable} in $\mathcal{A}$ relative to parameters $\bar{a}=(a_1,\dots,a_n)$ from $\mathcal{A}$ if there is a family of existential formulas $(\varphi_i)_{i\in\mathbb{N}}$ such that for all $x,y\in\mathcal{A}$, $f(x)=y$ if and only if there is an $i\in\mathbb{N}$ with $\mathcal{A}\models\varphi_i(\bar{a},x,y)$.

\noindent\textbf{Problem.} Is there a countable computably-represented structure $\mathcal{A}$ such that $\mathcal{A}$ is computably $\mathit{AUT}$-countable on a cone, and such that for any finite set of parameters $\bar{a}$ in $\mathcal{A}$ there is an automorphism $\pi$ of $\mathcal{A}$ which is not $\Sigma^{in}_1$-definable in $\mathcal{A}$ relative to the parameters $\bar{a}\cup\pi(\bar{a})$, where $\pi(\bar{a})$ is the image of $\bar{a}$ under $\pi$?

\begin{novelty}{Ted Slaman and Dariusz Kalociński}
A basic observation is that when the automorphism group of $\mathcal{A}$ is countable, there is a finite set of parameters in $\mathcal{A}$ such that every automorphism of $\mathcal{A}$ is determined by its action on that set. Intuitively, our question asks whether there is an $\mathcal{A}$ for which every automorphism can be effectively defined relative to some set of parameters in $\mathcal{A}$, but there is no single set of parameters such that every automorphism $\pi$ can be effectively defined relative to that set and its set of images under $\pi$.

As it turns out, there is such a structure.

There are many paradigms to produce structures with countable automorphism groups, and also standard diagonal methods to build structures while controlling which functions are definable from which parameters within those structures. The novelty of the solution that we present is that it does not apply any of those diagonal methods. Instead, the example presented is a remix of the additive group on the rational numbers, and the proof that it has the desired properties is geometrically motivated. Once the pattern of argument is formulated, the verification is elementary.

It took us several days, but not weeks, to find this example.\footnote{One of the referees pointed out that an expert who was sufficiently versed in the work of Turetsky and the work of  Alvir--Greenberg--Harrison-Trainor--Turetsky could solve the problem in a few hours.}

\end{novelty}

\subsection{Problem 2: Richard Schwartz (discrete geometry)}

Let $\mathbb{Z}$ denote the integers and let $\mathbb{R}$ denote the reals. Let $\Sigma=\mathbb{R}\times[0,1]$ and let $\partial\Sigma=\mathbb{R}\times\{0,1\}$ denote the boundary of $\Sigma$. For $\beta>0$ define $G_{\beta}\colon\Sigma\to\Sigma$ by
\[
G_{\beta}(x,y)=(x+\beta,\,1-y).
\]
Let $G_{\beta}^{k}$ denote the $k$-fold composition of either $G_{\beta}$ or its inverse $G_{\beta}^{-1}$, according as $k>0$ or $k<0$; when $k=0$, $G_{\beta}^{k}$ is the identity.

A \emph{clean triangle} is the convex hull of $3$ non-collinear points in $\partial\Sigma$. A \emph{clean triangulation} of $\Sigma$ is a countable and locally finite collection of clean triangles, having pairwise disjoint interiors, whose union is $\Sigma$. The clean triangulation is \emph{$G_{\beta}$-invariant} if the action of $G_{\beta}$ preserves the triangulation.

A \emph{squeeze map} of a clean triangle $\tau$ is an affine map from $\tau$ into $\mathbb{R}^3$ which decreases the length of the edge of $\tau$ that is contained in $\partial\Sigma$ and increases the lengths of the other two edges.

Call $\beta>0$ \emph{realized} if there is a continuous map $f\colon\Sigma\to\mathbb{R}^3$ with the following properties:
\begin{enumerate}[itemsep=1pt]
\item $f$ restricts to a squeeze map on each triangle of a $G_{\beta}$-invariant clean triangulation of $\Sigma$;
\item $f(p)=f(q)$ if and only if $p=G_{\beta}^{k}(q)$ for some $k\in\mathbb{Z}$ (for all $p,q\in\Sigma$).
\end{enumerate}
\noindent\textbf{Problem.} Prove that $\sqrt{3}$ is the infimum of the set of realized numbers.

\begin{novelty}{Richard Schwartz}
\textbf{Problem context.} The problem presented is a mathematical variant of the 1977 Halpern--Weaver Conjecture about paper Moebius bands. The question goes back at least to a 1962 paper by Wunderlich. I solved this conjecture several years ago (R.~E.~Schwartz, \textit{The optimal paper Moebius band}, Annals of Math., 2025). The main result in my paper is that if a flat Moebius band has a smooth isometric embedding into $\mathbb{R}^3$ then its aspect ratio is greater than $\sqrt 3$. It had already been known that there exist paper Moebius bands of aspect ratio $\sqrt 3+\epsilon$ for any $\epsilon>0$, so $\sqrt 3$ is the sharp result. My solution is kind of unusual: it was an old conjecture, but the solution uses only undergraduate (or maybe beginning graduate) level mathematics. The difficulty with the conjecture is that people did not know how to use the topological hypothesis in a geometry problem.

\textbf{Novelty.} The problem I am asking the AI is different from the one I solved in my paper in an important way. Here, I formulate the problem in terms of continuous piecewise affine maps which are not even local isometries. One virtue of this formulation is that the solution can be given entirely in terms of affine and Euclidean geometry; there is no reliance on results from the literature about zero-curvature smooth surfaces. It is all self-contained. The explicit connection to Moebius bands is that the quotient $\Sigma/G_\beta$ is a flat Moebius band. The problem is phrased so that the AI cannot use any of the results in my paper in an off-the-shelf way: it would either have to come up with its own proof, or else recognize that my paper is relevant, understand its proofs, and translate everything into a different category of maps.

\textbf{Proof strategy.} There are two halves. The first is showing that $\beta>\sqrt 3$. The central insight is to prove, using a variant of the Borsuk--Ulam Theorem, that there exists a pair of straight line segments $A,B\subset\Sigma$ such that:
\begin{itemize}
\item both $A$ and $B$ have their endpoints in $\partial\Sigma$;
\item the images $f(A)$ and $f(B)$ are perpendicular, coplanar, and have disjoint interiors;
\item the restriction of $f$ to each of $A$ and $B$ is a distance-increasing affine map.
\end{itemize}
Once this is in place, the idea (after $A$ and $B$ are suitably ordered) is to compare the geometry of the trapezoid bounded by $A$ and $G_\beta(A)$ with the constraints forced by $f(A)$ and $f(B)$; the comparison is a simple but kind of miraculous optimization problem in planar Euclidean geometry. The second half involves an explicit construction showing each $\beta=\sqrt 3+\epsilon$ can be realized, by correctly perturbing away from the so-called triangular Moebius band.

\textbf{Difficulty.} I would expect that a good graduate student (or a stellar undergraduate) in geometry and/or low-dimensional topology could solve this fairly quickly, knowing my solution to the Halpern--Weaver Conjecture. Without knowing my solution, one might work on this problem for years: the original problem was open for almost 50 years and I spent about 4 years (off and on) trying to solve it. If the AI finds a correct solution that does not resemble my own, I will be wildly impressed.
\end{novelty}

\subsection{Problem 3: Aleksa Milojevi\'c and Benny Sudakov (discrete probability)}

This problem and its solution will appear as part of the forthcoming paper:
\begin{itemize}
\item {\it Aleksa Milojevi\'c and Benny Sudakov, On the Probability a Weighted Bernoulli Sum Exceeds Its Mean, manuscript.}
\end{itemize}

Let $w_1,\dots,w_m$ be non-negative real numbers such that $\sum_{i=1}^{m}w_i=1$, and let $v_1,\dots,v_m$ be i.i.d.\ Bernoulli$(p)$ random variables. For which values of the probability $p$ is it true that
\[
\Pr\Big[\sum_{i=1}^{m}w_i v_i\geq p\Big]\geq p?
\]

\begin{novelty}{Aleksa Milojevi\'c and Benny Sudakov}
The main novelty of our argument is observing the connection of the probabilistic inequality $\Pr\big[\sum_{i=1}^m w_i v_i\geq p\big]\geq p$ with the Manickam--Mikl\'os--Singhi (MMS) conjecture. This conjecture states that for any positive integers $n,k$ with $n\geq 4k$, any set of $n$ real numbers with nonnegative sum has at least $\binom{n-1}{k-1}$ subsets of size $k$ with nonnegative sum. In particular, we observe that a solution to the MMS conjecture would imply this probabilistic inequality, and we use the results of Pokrovskiy resolving a weak form of the MMS to establish our inequality for $p\leq 10^{-46}$.

The reduction to the MMS conjecture is based on a coupling argument, which generalizes a previous argument of Alon, Emek, Feldman and Tennenholtz (which dealt only with the case $p=1/n$ for $n\in\mathbb{Z}_{>0}$). Although nontrivial, this coupling argument is not the most important element of our proof---observing the connection to the MMS conjecture is.

It is therefore hard to judge precisely how hard the problem is: beyond a nice idea, one also needs wide-area knowledge to spot the connection mentioned above. Still, it is safe to say that this observation qualifies more as a nice note than as a fundamental breakthrough. That being said, we suspect that proving the inequality for all $p\leq 1/3$ \footnote{$p\leq 1/3$ is the range in which one might suspect that inequality holds.} may be comparably hard as solving the MMS conjecture fully. This would be a very hard task, as witnessed by the fact that this almost 40-year old conjecture is still only partially resolved, despite a series of serious attempts.
\end{novelty}

\subsection{Problem 4: Larry Guth (metric geometry)}

Suppose $(M,g)$ and $(N,h)$ are piecewise smooth Riemannian manifolds. If $f\colon M\to N$ is piecewise smooth, we define the $k$-dilation of $f$ as the infimal $\Lambda$ so that for every $k$-dimensional submanifold $\Sigma\subset M$,
\[
\Vol_k(f(\Sigma))\le\Lambda\,\Vol_k(\Sigma).
\]
We write the $k$-dilation of $f$ as $\Dil_k(f)$. Equivalently, $\Dil_k(f)=\sup_{x\in M}\|\Lambda^k df\|$.

\noindent\textbf{Proposition.} Prove that there is a constant $k>0$ so that the following holds. Suppose that $R\subset\mathbb{R}^4$ is a $4$-dimensional rectangle with side lengths $R_1\le R_2\le R_3\le R_4$ and $S$ is a $4$-dimensional rectangle with side lengths $S_1\le S_2\le S_3\le S_4$. Suppose that $f\colon(R,\partial R)\to(S,\partial S)$ is a piecewise smooth map with degree $1$ and with $\Dil_2(f)\le 1$. Suppose that $R_1\le k S_1$. Then either
\[
R_3 R_4>k S_3 S_4,\qquad\text{or}\qquad R_1 R_2 R_3 R_4>k S_1 S_2^{1/2} S_3^{3/2} S_4.
\]

\begin{novelty}{Larry Guth}
The problem asks to study the 2-dilation of degree-1 maps between 4-dimensional rectangles. More precisely, if $R$ and $S$ are the two rectangles, we are asking about degree-1 maps $f:(R,\partial R)\to(S,\partial S)$. There is a simpler problem that was solved in all dimensions, which asks for a subset $U\subset R$ and a degree-1 map from $(U,\partial U)$ to $(S,\partial S)$. The main novelty is a new way to distinguish these two similar problems.

To control $f$, we consider rectangles $\Sigma\subset S$ and their inverse images $f^{-1}(\Sigma)$. We note that $f$ maps the complement of $f^{-1}(\Sigma)$ to the complement of $\Sigma$. These complements have additional topology. We can map the complement of $\Sigma$ to a wedge of spheres, and the resulting composition has non-trivial rational homotopy invariants. These can be defined using differential forms and related to the 2-dilation of $f$. The connection between rational homotopy invariants and differential forms is a technique in the literature, but the topology of this situation is a little different from previous ones. In this setting, the argument leads to a lower bound on the 3-volume of any chain with boundary $f^{-1}(\Sigma)$. Using the isoperimetric profile of $R$ (proven in previous work), this leads to a lower bound on the area of $f^{-1}(\Sigma)$, which leads to the given estimates.

The proposition came from an unpublished paper that I worked on as a postdoc. The paper had three somewhat novel steps, of which this is the first. This particular proposition probably took about two weeks of work, during which I spent about a third of my work time on it.
\end{novelty}

\subsection{Problem 5: 
Oleg Butkovsky,
Jonathan C. Mattingly, and 
Lorenzo Zambotti (stochastic PDE)}

This problem and its solution will appear as part of the forthcoming paper:
\begin{itemize}
\item 
{\it Oleg Butkovsky
Jonathan C. Mattingly 
Lorenzo Zambotti, manuscript in preparation, 2026.}
\end{itemize}

Let $u(x,t)$ be a real-valued solution to the following stochastic partial differential equation (SPDE):
\begin{align*}
    \frac{\partial u}{\partial t}(x,t) &= \frac{1}{2}\frac{\partial^2 u}{\partial x^2}(x,t) + \delta_0(u(x,t)) + \dot{W}(x,t)\quad\text{for }t\geq 0,\\
    u(x,0) &= u_0(x)\in C_0([0,1]),\\
    u(0,t) &= u(1,t)=0\quad\text{for all }t\geq 0,
\end{align*}
where $C_0([0,1]):=\{f\in C([0,1],\mathbb{R}):f(0)=f(1)=0\}$, $x\in[0,1]$, $t\geq 0$, $\delta_0$ is the Dirac delta function at $0$, $\dot{W}(x,t)$ is space-time white noise, and $u_0\in C_0([0,1])$ is the initial condition. We understand the probabilistically weak PDE mild solution of the equation in the sense described in Athreya, Butkovsky, Lê, and Mytnik, ``Well-posedness of stochastic heat equation with distributional drift and skew stochastic heat equation'' (2024) [ABLM24]. We assume that the existence, uniqueness, and approximation results of [ABLM24] extend to the SPDE on $[0,1]$ with Dirichlet boundary conditions, i.e.\ with the Dirichlet heat kernel and initial data $u_0\in C_0([0,1])$; in particular, that there exists a unique (in law) adapted mild solution on $C([0,T],C_0([0,1]))$.

We further assume that if $u^n$ is a solution to this SPDE with the drift $G_{1/n}$ in place of $\delta_0$, where $G_{1/n}(x):=\sqrt{\frac{n}{2\pi}}\exp(-\frac{n x^2}{2})$ is the 1D heat kernel on $\mathbb{R}$ with variance $1/n$, then for any $t>0$ we have $u^n(t)\to u(t)$ weakly as $n\to\infty$.

Let $P_t$ be the Markov semigroup associated with the solution $u(t)$, i.e.\ for any bounded measurable $f\colon C_0([0,1])\to\mathbb{R}$,
\[
P_t f(u_0)=\mathbb{E}\big[f(u(\cdot,t))\mid u(\cdot,0)=u_0(\cdot)\big],\quad u_0\in C_0([0,1]).
\]
\noindent\textbf{Problem.} Prove or disprove: $P_t$ has at most one invariant probability measure $\mu$ on $C_0([0,1])$.

\begin{novelty}{Jonathan Mattingly}

\textbf{Novelty.} There are a number of classical methods---Doob's Theorem using the Strong Feller property, Dirichlet form methods, or coupling using Girsanov's Theorem---that one might hope to apply to prove unique ergodicity. The difficulty is that the drift term, which contains a delta function at zero, is not regular enough to apply those methods; the regularization due to the Laplacian is not sufficient. We apply asymptotic Strong Feller / asymptotic coupling methods to prove the theorem. Those applications are not straightforward either, again due to the irregularity of the drift.

\textbf{Key references.} ``Skew heat equation'' (2011) by Bounebache--Zambotti shows the existence of the invariant measure but comments that proving the Strong Feller property seems out of reach. ``Well-posedness of stochastic heat equation with distributional drift and skew stochastic heat equation'' (2023) by Athreya, Butkovsky, L\^e, and Mytnik extends the results but still does not give the tools to prove uniqueness.  We needed to include some estimates from a stochastic sewing lemma, which not standard in this domain.

\textbf{Difficulty.} We first discussed this problem in October 2025 as a group, building on work over the preceding years by Butkovsky and Zambotti, and had the proof done by January 2026. In total it took 4-5 weeks. However, this also required the right mix of people: some had the ergodicity proof idea, while others knew how to prove the needed regularity estimates.
\end{novelty}

\subsection{Problem 6: Joshua Greene and Duncan McCoy (lattice theory)}

This problem and its solution will appear as part of the forthcoming paper:
\begin{itemize}
\item {\it Joshua Evan Greene and Duncan McCoy, On the Neumann-Zagier conjecture, 2026, in preparation.}
\end{itemize}
\bigskip

Let $T$ denote a finite, weighted tree with vertex set $V$, edge set $E$, and weight function $w\colon V\to\mathbb{Z}$. Let $L(T)$ denote the free abelian group generated by $V$, equipped with the symmetric bilinear form $\cdot$ whose value on a pair of generators $u,v\in V$ is given by
\[
u\cdot v=
\begin{cases}
w(u), & u=v;\\
-1, & (u,v)\in E;\\
0, & u\ne v,\ (u,v)\notin E.
\end{cases}
\]
Suppose that the form is positive definite, and suppose that there exists a single vertex $v$ for which the vertex weight $w(v)$ is less than the vertex degree $d(v)$. Prove that $T$ contains a vertex $u$ such that $u$ is an irreducible element of $L(T)$, i.e.\ there do not exist nonzero lattice elements $a,b\in L(T)$ such that $u=a+b$ and $a\cdot b\ge 0$.

\begin{novelty}{Joshua Evan Greene and Duncan McCoy}
\begin{itemize}
\item[(i)] We believe the argument is rather novel. There is not much literature on the Neumann--Zagier conjecture. The closest related paper is one of Issa and McCoy; however, our argument is very different from the argument in that paper. There are several subtle steps in our solution which depart from a straightforward attempt.
\item[(ii)] The general strategy of the proof is to argue that either the bad vertex is irreducible, or else there is a leaf of weight one (which is therefore irreducible). However, we make more delicate arguments and estimates, drawing on properties of the dual lattice: some known, some novel. Our submitted solution is laid out somewhat in reverse. Lemma~2.6 shows that if a vertex in a definite tree lattice is reducible, then some associated dual element is either reducible or else is contained in the lattice itself. We apply this to the bad vertex, the one whose weight is lower than its degree. Corollary~2.3 rules out the possibility that any associated dual element is irreducible; this is a quite novel argument, which sensitively uses the tree structure, and which backs onto an estimate for the lengths of elements in dual cut lattices, going back to Kirchhoff. Lemma~2.2 rules out the possibility that any associated dual element is contained in the lattice itself, unless there is a leaf of weight one.
\item[(iii)] We made an attempt on the Neumann--Zagier conjecture over the course of Summer 2020, getting the special case in which the tree lattice is isometric to the standard Euclidean lattice. We made another effort over a few weeks during Fall 2023, in which we realized we could prove the stronger result. All told, it was a serious effort by both of us over the course of several (4--5) months.
\end{itemize}
\end{novelty}

\subsection{Problem 7: Sucharit Sarkar (combinatorial topology)}

This problem and its solution will appear as part of the forthcoming paper:
\begin{itemize}
\item
{\it Robert Lipshitz, Lenhard Ng, and Sucharit Sarkar, A stable homotopy refinement of Legendrian contact homology, in preparation.}
\end{itemize}
\bigskip

Let $G=\langle a_1,\dots,a_r\rangle$ be a free group. A word on the letters $L=\{a_1,a_1^{-1},\dots,a_r,a_r^{-1}\}$ is called \emph{reducible} if it equals $1\in G$. A word is called \emph{quasi-reduced} if it does not contain three consecutive letters $\ell\ell^{-1}\ell$ for any $\ell\in L$.

Fix a reducible quasi-reduced word $w=w_1\cdots w_{2n}$ (with $w_i\in L$). A \emph{matching} $M$ consists of $n$ intervals, each properly embedded in $\mathbb{R}\times\mathbb{R}_+$, satisfying:
\begin{itemize}[leftmargin=*,itemsep=1pt]
\item any two intervals intersect transversally, and there are no triple points;
\item $\partial M=\{1,\dots,2n\}\times 0\subset\mathbb{R}\times 0\subset\mathbb{R}\times\mathbb{R}_+$;
\item for each component $R$ of $\mathbb{R}\times\mathbb{R}_+\setminus M$, and for each component $A$ of $M\cap\partial\overline{R}$, $\partial A$ consists of some two points $(i,0)$ and $(j,0)$ with $w_i=w_j^{-1}$.
\end{itemize}
A \emph{$k$-crossing matching} is a matching where the total number of self-intersections is $k$.

Let $F_w$ be the following CW complex whose $k$-cells correspond to isotopy classes of $k$-crossing matchings. Given a $k$-crossing matching $M$ with crossing set $X(M)$, the associated $k$-cell is $C(M):=\prod_{c\in X(M)}I_c$, where $I_c$ is an interval whose endpoints correspond to the two ways of locally resolving $M$ near $c$. The attaching maps are as follows: on a facet of the form
\[
a\times\prod_{d\in X(M)\setminus\{c\}}I_d,\qquad c\in X(M),\ a\in\partial I_c,
\]
let $M_a$ be the $(k-1)$-crossing matching obtained by resolving $M$ at $c$ according to $a$. Then the $(k-1)$-cell associated to $M_a$ is $C(M_a)=\prod_{d\in X(M_a)}I_d=\prod_{d\in X(M)\setminus\{c\}}I_d$, and the above facet maps to this $(k-1)$-cell canonically.

\noindent\textbf{Problem.} Is $F_w$ contractible?

\begin{novelty}{Sucharit Sarkar}
The novelty lies in the definition of the CW complex $F_w$. While similar definitions have appeared elsewhere (like matching complexes), this specific definition of $F_w$ appears to be completely new. (In the actual paper, we study an even more general complex, where the restrictions ``quasi-reduced'' and ``reducible'' on $w$ are dropped.) The challenge is in ``comprehending'' this new concept. Once one understands what the definition actually represents, the properties of $F_w$ are fairly easy to deduce. For instance, $F_w$ is always $1$-connected---it took the authors around $1$ hour to prove that; and then they tried desperately to prove that $F_w$ is $2$-connected as well. After around 2--3 hours of unsuccessfully trying to adapt the previous proof, they tried to find a counterexample, which they then immediately found (much to their disappointment).
\end{novelty}

\subsection{Problem 8: Sam Payne and Jidong (Jayden) Wang (matroids and tropical geometry)}

This problem and its solution will appear as part of the forthcoming paper:
\begin{itemize}
\item
{\it
Jidong Wang, in preparation, 2026.}
\end{itemize}

Let $M$ be a matroid, and let $\Trop(M)$ be the corresponding tropical linear space, i.e.\ the closure of the Bergman fan of $M$ in tropical projective space. The \emph{relative Dressian} of $M$ is
\[
\Dr(M):=\{\text{tropical linear subspaces of }\Trop(M)\},
\]
which is partially ordered by inclusion of tropical linear spaces. Suppose $M$ admits an order-reversing involution $F\mapsto F^\perp$ on the lattice of flats $\mathcal{L}(M)$. Note that there is a natural inclusion $\mathcal{L}(M)^{\mathrm{op}}\subset\Dr(M)$ given by $F\mapsto\Trop(M/F\oplus U_{0,F})$.

\noindent\textbf{Problem.} Is there an order-reversing involution $\Phi$ on $\Dr(M)$ that extends the involution $F\mapsto F^\perp$?

\begin{novelty}{Sam Payne and Jidong (Jayden) Wang}
This problem is at the level of an interesting technical proposition. We expect that an advanced graduate student working on matroid theory and tropical linear spaces would be able to find a solution. The proof requires three steps: (1) understanding that the answer should be affirmative; (2) constructing the map, either directly as a tropical linear map on the ambient tropical projective spaces containing the corank-$k$ Dressians in their respective Pl\"ucker embeddings, or by passing to a cryptomorphic characterization of valuated quotients as functions on lattices (as in our solution); (3) verifying that this map is well-defined, i.e.\ that it takes points in the corank-$k$ Dressian to points in the rank-$k$ Dressian.

We estimate that a sufficiently advanced AI system, by current standards, should solve (1) and (2) with probability 60--70\%, but giving a clear and correct verification (step (3)) may be more challenging, with a probability of success of 40--50\%.
\end{novelty}

\subsection{Problem 9: Sylvie Corteel and John Lentfer (algebraic combinatorics)}

This problem and its solution will appear as part of the forthcoming paper:
\begin{itemize}
\item {\it Sylvie Corteel and John Lentfer, Universal Hilbert series coefficients of the superspace coinvariant ring, 2026, in preparation.}
\end{itemize}

\bigskip

Let $m$ and $n$ be positive integers and let $R_n^{(m)}$ be the coinvariant algebra with $m$ sets of $n$ variables. The Hilbert series of an $m$-multigraded algebra $A$ is
\[
\mathrm{Hilb}(A;q_1,\dots,q_m)=\sum_{i_1,\dots,i_m\geq 0}\dim\big(A_{(i_1,\dots,i_m)}\big)\,q_1^{i_1}\cdots q_m^{i_m},
\]
where $A_{(i_1,\dots,i_m)}$ denotes a multigraded component, with grading variables $q_1,\dots,q_m$. Let $s_\lambda(q_1,\dots,q_m)$ denote a Schur polynomial. For $R_n^{(m)}$, it is known that its Hilbert series may be expressed by
\[
\mathrm{Hilb}(R_n^{(m)};q_1,\dots,q_m)=\sum_{\lambda}c_\lambda(n)\,s_\lambda(q_1,\dots,q_m),
\]
for some nonnegative universal series coefficients $c_\lambda(n)$ which do not depend on $m$.

\noindent\textbf{Problem.} For any $n$, give a combinatorial interpretation of the coefficients $c_\lambda(n)$ when $\lambda$ is a hook shape partition.

\begin{novelty}{Sylvie Corteel and John Lentfer}
\textbf{Novelty.} The definition of $R_n^{(m)}$ implies that if you know its Hilbert series and compute the Schur expansion, you get the coefficient $c_\lambda(n)$ for all partitions $\lambda$ with at most $m$ parts. For now this is known for $m=1$, and the Hilbert series is known for $m=2$ but not its Schur expansion. The only known hook shape is $\lambda=(a,1^b)$ with $b=0$; Opus, Gemini and ChatGPT know this result.

\textbf{Strategy of the proof.} The ``trick'' here is due to a result of John Lentfer during his PhD at Berkeley: if you compute the Hilbert series of the coinvariant algebras in $m$ sets of commuting variables and $j$ sets of anticommuting variables, and know their super Schur expansion, then you get the coefficient $c_\lambda(n)$ for any $n$ and any $\lambda$ with $\lambda_{m+1}\le j$.

\textbf{Difficulty.} Once you know the trick, the hook shape corresponds to $m=j=1$. The proof then uses a sign-reversing involution of Sagan and Swanson together with super Schur polynomials. Once you mix all these ingredients, the proof is not too long.
\end{novelty}

\subsection{Problem 10: Srivatsav Kunnawalkam Elayavalli
(von Neumann algebras)}

The solution of this problem
(which appears in the auxiliary solutions document) was written by
\begin{itemize}
\item {\it Srivatsav Kunnawalkam Elayavalli
and Zhiyuan Yang}
\end{itemize}

\noindent\textbf{Problem.} Let $\Gamma$ be a finite simplicial graph with $|V(\Gamma)|\geq 3$, and let $(M_v,\varphi_v)$ be a separable tracial von Neumann algebra for each $v\in V$. If $\Gamma$ is irreducible (in the sense that there do not exist subgraphs $\Gamma_1,\Gamma_2$ such that $V(\Gamma_1)\cup V(\Gamma_2)=V(\Gamma)$ and $(v,u)\in E(\Gamma)$ for all $v\in V(\Gamma_1),u\in V(\Gamma_2)$), and each $M_v$ has a trace-zero unitary, then the graph product von Neumann algebra $M$ is properly proximal.

\begin{novelty}{Srivatsav Kunnawalkam Elayavalli and Zhiyuan Yang}
\textbf{Novelty of the argument.} The argument is a von Neumann algebraic analogue of the argument in the case of graph products of groups, as executed in the proof of Theorem~1.8 of the work of Ding and Kunnawalkam Elayavalli (2024). However, the implementation is difficult, because it has to be done in the intricate language of small-at-infinity compactifications of von Neumann algebras developed by Ding, Kunnawalkam Elayavalli and Peterson. Various subtleties arise in this setting that require nontrivial arguments: a fact well known to experts is that not every group-theoretic argument generalizes immediately to von Neumann algebras. For instance, the results of Popa and Johnson--Parrot rule out the existence of derivations of a von Neumann algebra into the compact operators, despite the fact that plenty of these exist in the group-theoretic analogue.

\textbf{General strategy of proof.} The proof follows a ``patching'' argument, available thanks to the decompositional structure of graph products. Knowing that the graph is irreducible allows us to split the graph product into various nontrivial amalgamated free products, where each amalgam is the star of a fixed vertex. This yields ``relative proper proximality'' of the entire algebra with respect to the collection of subalgebras generated by the star subgraphs---an extension of the paradoxical decomposition trick that proves proper proximality for free products. The next part is more subtle, and involves upgrading this relative proper proximality, by proving that the support projections of the associated boundary pieces commute with each other and that their infimum is the compact operators. This is proved by combinatorially exploiting the Fock space representation of graph products (Caspers--Fima).

\textbf{Estimate of difficulty.} Being able to carry out a proof of this result would be a good marker for a strong third- or fourth-year graduate student working in this subfield of von Neumann algebras. I expect they would take a couple of weeks (two or three) to nail down and write up the argument in full detail.
\end{novelty}

\section{AI Systems Tested}
\label{sec:systems}
The four systems below are listed in the order that their participation in the benchmark was confirmed.

\subsection{System A: IMProofBench ProofCouncil}
System A, called ProofCouncil, was developed by a team consisting of
Johannes Schmitt (ETH Zürich), Tim Gehrunger (ETH Zürich), Jasper Dekoninck
(ETH Zürich), Gergely Bérczi (Aarhus University), Uri Kreitner (ETH Zürich),
and Liam Price (Independent Researcher). The system primarily used
\texttt{gpt-5.5 pro} as its base model, though it also called
\texttt{gpt-5.5}, \texttt{gemini-3.1-pro-preview}, and
\texttt{claude-opus-4-7}. The source code for System A is available at
\href{https://github.com/1stproof/batch-2/tree/main/batch-2-submissions/improofbench}{\texttt{batch-2-submissions/improofbench}}.

\subsection{System B: UCLA Moonshot Harness}
System B, called the UCLA Moonshot Harness, was developed by
Junyi Zhang\footnote{Co-first authors with equal contribution. The remaining students are ordered by contribution to the harness.}\newcounter{ucbcofirst}\setcounter{ucbcofirst}{\value{footnote}},
Xinjie He\footnotemark[\value{ucbcofirst}], Hyunsik Chae,
Ethan Ji, Eric Jiang, Rushil Raghavan, Yiwen Kou, Alex Taylor,
Kai-Wei Chang\footnote{Principal investigators, listed at the end in alphabetical order by last name.}\newcounter{ucbpi}\setcounter{ucbpi}{\value{footnote}},
Raghu Meka\footnotemark[\value{ucbpi}], Violet Peng\footnotemark[\value{ucbpi}],
Amit Sahai\footnotemark[\value{ucbpi}], Terence Tao\footnotemark[\value{ucbpi}], and
Wei Wang\footnotemark[\value{ucbpi}], all of the
University of California, Los Angeles.   The system used  \texttt{gpt-5.5 pro} as its base model.  The source code for System B is available at
\href{https://github.com/1stproof/batch-2/tree/main/batch-2-submissions/ucla}{\texttt{batch-2-submissions/ucla}}.
\subsection{System C: OpenAI ChatGPT 5.5 Pro}
System C is OpenAI's ChatGPT 5.5 Pro; the team includes Sébastien Bubeck
and Mehtaab Sawhney (OpenAI), with Sébastien Bubeck as the point of contact
for the benchmark. The following prompt was provided to First Proof by Mehtaab Sawhney. For all 10 problems, we used this prompt and we set the reasoning/thinking mode to 'xhigh' (the highest option). 

\bigskip

\begin{quote}
    
{\bf Prompt:} Consider the following research-level math question.

[Question]

Carefully consider the question and provide a complete and rigorous proof. Provide either complete proofs or careful citations for all intermediate statements in the proof. Work very hard to prove the question and do not return until a complete proof is achieved.

Make sure to structure the proof with careful lemmas and theorem statements. The output must be a compilable LaTeX document conforming to the standards of rigor and scholarship prevailing in mathematical literature.
\end{quote}
\subsection{System D: Princeton Momus}
System D, called Momus, was developed by Sanjeev Arora and 
Liam Fowl of Princeton University, and used \texttt{gemini-3.1-pro-preview} as its base model.
The source code for System D is available at
\href{https://github.com/1stproof/batch-2/tree/main/batch-2-submissions/princeton}{\texttt{batch-2-submissions/princeton}}. 

The Momus team reports that at least four calls show signs of an unforeseen timeout or race condition that corrupted the runs, one sign of which is outputs with "Disclaimer" statements inserted by the orchestrator. They regret that these unforeseen conditions tripped up the run.

\section{Results and Costs}\label{sec:results}
The results of the refereeing process are summarized in Table \ref{tab:decisiongrid}. Table \ref{tab:costs} shows the costs as well as the token usage for each system, broken down by problem. These costs do not include the costs of the AWS server required for these runs, which was minimal (less than \$35 for all the runs). 

\definecolor{essgreen}{RGB}{60,160,70}
\definecolor{minorange}{RGB}{235,150,40}
\definecolor{majred}{RGB}{200,40,40}
\definecolor{rejblack}{RGB}{40,40,40}
\newcommand{\rb}[1]{\tikz[baseline=-0.45ex]{\node[fill=#1,rounded corners=1.6pt,%
  minimum width=1.7ex,minimum height=1.7ex,inner sep=0pt]{};}\kern1.4pt}
\newcommand{\Ess}{\rb{essgreen}}
\newcommand{\Min}{\rb{minorange}}
\newcommand{\Maj}{\rb{majred}}
\newcommand{\Rej}{\rb{rejblack}}
\newcommand{\splitpill}[1]{\tikz[baseline=-0.45ex]{%
  \begin{scope}
  \clip[rounded corners=1.6pt] (-0.85ex,-0.85ex) rectangle (0.85ex,0.85ex);
  \fill[minorange] (-0.85ex,-0.85ex) -- (0.85ex,-0.85ex) -- (-0.85ex,0.85ex) -- cycle;
  \fill[#1]        (0.85ex,-0.85ex) -- (0.85ex,0.85ex) -- (-0.85ex,0.85ex) -- cycle;
  \end{scope}}\kern1.4pt}
\newcommand{\MinC}{\splitpill{essgreen}} 
\newcommand{\MinM}{\splitpill{majred}}   
\newcommand{\cell}[2]{{\hypersetup{hidelinks}\hyperref[#1]{#2}}}

\makeatletter
\Hy@raisedlink{\hyper@anchorstart{table.2}\hyper@anchorend}
\makeatother
\setcounter{table}{2}
\begin{table}[ht]
\centering
\renewcommand{\arraystretch}{1.7}
\setlength{\tabcolsep}{10pt}
\begin{tabular}{c cccc}
\toprule
 & System A & System B & System C & System D \\
\midrule
P1  & \cell{res:P1A}{\Ess\MinC}     & \cell{res:P1B}{\Ess\MinC\MinC} & \cell{res:P1C}{\Ess\MinC}     & \cell{res:P1D}{\Rej\Rej}      \\
P2  & \cell{res:P2A}{\MinC\MinC}    & \cell{res:P2B}{\MinM\MinC\MinC}& \cell{res:P2C}{\MinM\MinM}    & \cell{res:P2D}{\Rej\Rej\Rej}  \\
P3  & \cell{res:P3A}{\MinC\MinC\MinC}& \cell{res:P3B}{\Rej\Rej\Rej}   & \cell{res:P3C}{\Rej\Rej}      & \cell{res:P3D}{\Rej\Rej}      \\
P4  & \cell{res:P4A}{\Maj\Rej}      & \cell{res:P4B}{\Rej\Rej\Rej}   & \cell{res:P4C}{\Rej\Rej}      & \cell{res:P4D}{\Rej\Rej\Rej}  \\
P5  & \cell{res:P5A}{\Ess\Ess\Ess}  & \cell{res:P5B}{\Rej\Rej\Rej}   & \cell{res:P5C}{\Rej\Rej\Rej}  & \cell{res:P5D}{\Rej\Rej}      \\
P6  & {\color{gray}\footnotesize ---} & \cell{res:P6B}{\Ess\Ess\Ess} & \cell{res:P6C}{\Ess\Ess\Ess}  & \cell{res:P6D}{\Rej\Rej\Rej}  \\
P7  & \cell{res:P7A}{\Ess\Ess\MinM} & \cell{res:P7B}{\MinC\MinM\Maj} & \cell{res:P7C}{\Ess\MinC\MinM}& \cell{res:P7D}{\Ess\MinC\MinM}\\
P8  & \cell{res:P8A}{\Maj\Maj\Rej}  & \cell{res:P8B}{\Maj\Maj}      & \cell{res:P8C}{\Rej\Rej\Rej}  & \cell{res:P8D}{\Rej\Rej}      \\
P9  & \cell{res:P9A}{\MinC\MinM}    & \cell{res:P9B}{\MinC\MinM\MinM}& \cell{res:P9C}{\MinM\MinC}    & \cell{res:P9D}{\Rej\Rej\Rej}  \\
P10 & \cell{res:P10A}{\Maj\Maj}     & \cell{res:P10B}{\Rej\Rej}      & \cell{res:P10C}{\Maj\Maj\Maj} & \cell{res:P10D}{\Rej\Rej}     \\
\bottomrule
\end{tabular}
\caption[Referee recommendations per submission.]{Referee recommendations per submission:
{\Ess}~essentially flawless, {\Min}~minor revisions, {\Maj}~major revisions,
{\Rej}~reject. Minor-revision pills are split diagonally to show the basis for
the revision: {\MinC}~the mathematics is correct but the citations are inadequate and/or incorrect,  {\MinM}~the reviewer flags an actual
mathematical error or gap that must be corrected (whether in their summary or
in their inline review comments). Each cell is a link to the corresponding editor
summary of the referee reports for that problem. The entry for Problem~6,
Submission~A is blank because of a technical issue (see Section \ref{sec:P6}).}
\label{tab:decisiongrid}
\end{table}

\begin{table}[ht]
\centering
\footnotesize
\setlength{\tabcolsep}{4pt}
\resizebox{\textwidth}{!}{%
\begin{tabular}{l rrr rrr rrr rrr}
\toprule
 & \multicolumn{3}{c}{System A} & \multicolumn{3}{c}{System B} & \multicolumn{3}{c}{System C} & \multicolumn{3}{c}{System D} \\
\cmidrule(lr){2-4}\cmidrule(lr){5-7}\cmidrule(lr){8-10}\cmidrule(lr){11-13}
Prob. & In & Out & \$ & In & Out & \$ & In & Out & \$ & In & Out & \$ \\
\midrule
P1 & 0.22 & 0.24 & 49 & 1.25 & 0.84 & 189 & 0.11 & 0.05 & 13 & 6.03 & 10.84 & 142 \\
P2 & 0.46 & 0.40 & 86 & 3.14 & 2.60 & 562 & 0.10 & 0.05 & 12 & 3.37 & 6.54 & 85 \\
P3 & 105.49 & 4.91 & 712 & 4.32 & 2.82 & 637 & 0.15 & 0.06 & 16 & 2.85 & 6.17 & 80 \\
P4 & 115.73 & 6.43 & 951 & 1.87 & 0.85 & 208 & 0.13 & 0.03 & 10 & 3.24 & 6.01 & 79 \\
P5 & 14.21 & 1.61 & 313 & 6.12 & 2.55 & 642 & 0.11 & 0.04 & 11 & 3.07 & 4.60 & 61 \\
P6 & -- & -- & -- & 1.59 & 0.94 & 218 & 0.00 & 0.06 & 11 & 2.62 & 5.53 & 72 \\
P7 & 0.16 & 0.22 & 44 & 2.59 & 2.24 & 480 & 0.06 & 0.06 & 13 & 2.92 & 9.12 & 115 \\
P8 & 29.18 & 1.85 & 321 & 5.49 & 3.29 & 757 & 0.13 & 0.06 & 14 & 6.11 & 15.50 & 198 \\
P9 & 0.34 & 0.14 & 36 & 6.15 & 1.84 & 516 & 0.11 & 0.03 & 8 & 2.52 & 6.36 & 81 \\
P10 & 23.43 & 3.15 & 673 & 6.96 & 2.11 & 589 & 0.13 & 0.02 & 8 & 4.20 & 7.67 & 100 \\
\midrule
\textbf{Total} & 289.21 & 18.94 & 3186 & 39.47 & 20.08 & 4799 & 1.05 & 0.47 & 117 & 36.94 & 78.35 & 1014 \\
\midrule
\textbf{Wall clock} & \multicolumn{3}{c}{22.9\,h} & \multicolumn{3}{c}{23.1\,h} & \multicolumn{3}{c}{5.8\,h\footnotemark} & \multicolumn{3}{c}{7.8\,h} \\
\bottomrule
\end{tabular}%
}
\caption{Per-problem token usage and dollar cost for each submission. \emph{In} and
\emph{Out} are input and output tokens in millions; \emph{Out} includes
reasoning/thinking tokens; \$ is US dollars. The base system for Systems~A and B is \texttt{gpt-5.5-pro}, while System~D relies on \texttt{gemini-3.1-pro-preview}.
 System~A also makes limited use of other models: \texttt{gpt-5.5},
\texttt{gemini-3.1-pro-preview}, and \texttt{claude-opus-4-7}. Systems~A,
B, and~C costs are vendor-billed; \textbf{System~D costs are
imputed} by pricing its logged tokens at published
\texttt{gemini-3.1-pro-preview} list rates (\$2.00/M input, \$12.00/M output),
as its logs record tokens only. System~A's log is missing Problem~6
(\emph{---}); its large input counts are mostly cached tokens, billed far below
the fresh-input rate. \emph{Wall clock} is total elapsed run time: for
Systems~A, B, and~D the harness instance runtime, with problems run
concurrently.}
\label{tab:costs}
\end{table}
\footnotetext{System~C's wall clock is the serial sum of per-problem API
latencies over the nine problems whose latency was logged; Problem~6 was
obtained on a separate retry whose elapsed time was not recorded.}

\begin{table}[ht]
\centering
\renewcommand{\arraystretch}{1.4}
\setlength{\tabcolsep}{5pt}
\begin{tabular}{c cccccccccc}
\toprule
Sys.\ & P1 & P2 & P3 & P4 & P5 & P6 & P7 & P8 & P9 & P10 \\
\midrule
A & ---     & missing  & missing   & --- & missing & ---     & --- & incorrect & missing & incorrect \\
B & missing & missing  & incorrect & --- & missing & ---     & --- & ---       & both    & ---       \\
C & missing & incorrect& incorrect & --- & both    & missing & --- & incorrect & both    & incorrect \\
D & ---     & ---      & missing   & --- & missing & ---     & --- & ---       & both    & incorrect \\
\bottomrule
\end{tabular}
\caption{Reference and attribution problems flagged by the referees, per
submission. \emph{missing}: a relevant work that should have been cited was
not; \emph{incorrect}: a citation was given but is wrong, non-supporting, or
hallucinated; \emph{both}: both kinds were flagged; \emph{---}: none.
Submission~6A is blank.}
\label{tab:refproblems}
\end{table}

\subsection{Problem 1}
\paragraph{Submission A.}\label{res:P1A}
\editorialdecision{Minor Revisions.}
\refereereportlink{01}{1}{A}

One reviewer felt that the submission required {minor revisions} -- 
"the solution is correct and there are essentially only expository
concerns. There are some statements that are hard to understand
due to their wording, but they can generally be understood with some effort."
The other reviewer felt that the solution was {essentially flawless}.  They noted
``The only issue that I noticed, which is not serious, is that in the middle of the solution there is a high-level description of the strategy to complete the proof which is at best misleading and at worst simply a false description of what’s going on. However, this statement does not play a technical
role in the proof and all the technical details are correct.''

The reviewers pointed out that the ideas used in this solution were very similar to those used in the human solution.

\paragraph{Submission B.}\label{res:P1B}

\editorialdecision{Minor Revisions.}
\refereereportlink{01}{1}{B}

All three reviewers felt that the 
 solution required {minor revisions},  because of 
 egregious \textbf{missing citations}.
 In particular, all reviewers pointed out that there are similar arguments in the literature which need to be cited. 
"As I wrote in my review of another, similar, AI solution:
\begin{quote}
{\it If the AI system was aware of---and especially if it was directly inspired by---Turetsky's work as discussed above, then its failure to cite or even mention Turetsky (or any of the papers which use his technique) is a serious and very concerning oversight. It seems reasonably likely that this occurred,  especially given the overlap of the keywords in the problem and in the title of the paper of Turetsky "Coding in the Automorphism \dots".}
\end{quote}
The article should have cited 
references such as the following:
\begin{itemize}
\item Coding in the Automorphism Group of a Computably Categorical Structure, by Turetsky
\item Degrees of Categoricity and Treeable Degrees, by Csima and Rossegger
\item Scott Complexity of Countable Structures, by Alvir, Greenberg, Harrison-Trainor, and Turetsky
\end{itemize}

The solution given was similar to the solution given by Submission C.

In terms of exposition, the solution was sometimes so pedantic that it was hard for referees to follow the train of thought.  And the overview of the proof provided in the solution was unhelpful.  The solution also inserted proofs of statements that no mathematician would have bothered doing, and  introduced some unnecessary notation.

\paragraph{Submission C.}\label{res:P1C}
\editorialdecision{Minor Revisions.}
\refereereportlink{01}{1}{C}

Both reviewers agreed that this solution required {minor revisions} based on rather egregious \textbf{lack of citations}.  In particular, 
\begin{quote} if the AI system was aware of—and especially
if it was directly inspired by—Turetsky’s work, then its failure to cite or even mention Turetsky (or any of the papers which use his technique) is a serious and very concerning oversight. It seems reasonably likely that this occurred, especially given the overlap of the keywords in the problem and in the title of the paper of Turetsky "Coding in the Automosphism \dots".
\end{quote}

The article should have cited 
references such as the following:
\begin{itemize}
\item Coding in the Automorphism Group of a Computably Categorical Structure, by Turetsky
\item Degrees of Categoricity and Treeable Degrees, by Csima and Rossegger
\item Scott Complexity of Countable Structures, by Alvir, Greenberg, Harrison-Trainor, and Turetsky
\end{itemize}
The solution given was similar to the solution given by Submission B.

Apart from the  lack of citations, the solution was deemed to have good exposition, the best among the four AI-generated solutions (and perhaps slightly above average when compared to typical human solutions).  

\paragraph{Submission D.}\label{res:P1D}
\editorialdecision{Reject.} 
\refereereportlink{01}{1}{D}

Both reviewers agreed that this submission should be {rejected}.  The 
submission attempts to prove a false statement. The answer to the problem is positive, and the submission attempts to prove it is negative.
The solution added a hypothesis it didn't need, and spent most of the time verifying the unnecessary hypothesis.

\subsection{Problem 2}

None of these solutions included figures and diagrams, which are necessary in this subject for mathematicians to be able to effectively understand the correctness of the constructions. All solutions followed the outline of a recent paper by  Schwartz \cite{Schwartz}. 

\paragraph{Submission A.}\label{res:P2A}

\editorialdecision{Minor Revisions.}
\refereereportlink{02}{2}{A}

The first half of this submission lacked appropriate attributions: it adapts Schwartz's paper~\cite{Schwartz} in some parts line by line, reusing its terminology (``$T$-patterns,'' ``bends'') and even its labels $B,T,D,H$, without citing it anywhere. Both reviewers flagged this, and this would likely lead to an immediate rejection if it were a human journal submission. The second half of this solution included the construction of a six-triangle realization that was surprising to the reviewers. Given the fact that no figures were included, the AI system produced a surprisingly comprehensible notation and proof, with which the reviewers were impressed once they disentangled it.

\paragraph{Submission B.}\label{res:P2B}

\editorialdecision{Minor Revisions.}
\refereereportlink{02}{2}{B}

This solution is correct. It nonetheless showed an unacceptable level of uncredited mathematical ideas and notation: the lower bound reproduces Schwartz's arguments in lightly disguised form, again down to the labels $T,B,D,H$, and while its bibliography cites Sadowsky~\cite{Sadowsky} and Wunderlich~\cite{Wunderlich}, it omits Schwartz's paper~\cite{Schwartz} altogether. As one reviewer whose identity can be deduced from the review wrote: ``given that these other papers are cited, it is pretty shocking that mine is not.'' In addition, the proof of the second part of the solution is inelegant and lacks conceptual explanation for the computations, invoking the smooth Sadowsky construction as a black box and grinding through a PL approximation. One reviewer noted an unclear step, which upon further analysis ends up immediately following from a formula that can be found in some linear algebra textbooks.

\paragraph{Submission C.}\label{res:P2C}

\editorialdecision{Minor Revisions.}
\refereereportlink{02}{2}{C}

This submission followed the same strategy as Submission B, but the writing was a little bit better. It does cite Schwartz's paper~\cite{Schwartz}, but only as background for the second half, without acknowledging that the first half closely follows him. In addition, there is an incorrectly cited result for one of the arguments: it attributes a smooth isometric embedding to Sadowsky~\cite{Sadowsky}, whose construction is only $C^1$. The first smooth embedding is due to Halpern--Weaver (1977)~\cite{HalpernWeaver}. The reviewers decided not to use this to demote the solution to a lower category, because it would be easy to fix the references given the provided comments.

\paragraph{Submission D.}\label{res:P2D}
\editorialdecision{Reject.}
\refereereportlink{02}{2}{D}

The attempts made in the first section were incorrect, and the model itself conceded that the solution was only partial. For the lower bound it appeals directly to Schwartz's theorem~\cite{Schwartz} as though ``any embedded ruled Moebius band'' has aspect ratio $\ge\sqrt3$, but that is essentially the statement to be proved, and it does not apply here, since the squeeze maps are not isometric on the triangles. There was not enough detail in the second part to qualify this as a mathematical solution: the realization half is a vague heuristic with no explicit construction.

\subsection{Problem 3}

\paragraph{Submission A.}\label{res:P3A}
\editorialdecision{Minor Revisions.}
\refereereportlink{03}{3}{A}

The submission establishes counterexamples for $p$ in $(1/3,1/2) \cup (1/2,1)$, and proves the inequality for $p= 1/k$, $k \in \mathbb{Z}_{+}$.  The submission also establishes solutions for the family $p = 2/k$, $k \geq 6$. This last family, obtained via a pairing lemma, appears to be novel.  The central tool is a coloring criterion which, after a trivial renormalization, is an instance of the Manickam-Miklos-Singhi (MMS) conjecture.  

The submission did not make the connection between the coloring criterion and the MMS conjecture.  If it had, it could have then discovered the result of Pokrovskiy which would have established the condition for $p \leq 10^{-46}$. 

All three reviewers recommended minor revisions; the noted weaknesses were the missing MMS connection as well as missing citations; the mathematical content was  judged correct and genuinely novel.

\paragraph{Submission B.}\label{res:P3B}
\editorialdecision{Reject.}
\refereereportlink{03}{3}{B}

The proof rests on a false claim: the comparison theorem (Lemma 3), for which the reviews construct an explicit counterexample. This lemma and a second key input (Lemma 4, a binomial mean-tail estimate) are both represented as consequences of the cited literature, but neither result seems to appear in the cited sources, and no valid derivation is supplied. The submission does correctly establish that the inequality fails universally for every
$p$ outside $[0, 1/3] \cup \{1/2, 1\}$ and correctly handles the elementary cases $p = 1/2$ and $p = 1$.  However, the reduction itself is already standard, and the central connection to the Manickam-Miklos-Singhi conjecture is never made. All three reviewers recommend rejection.

\paragraph{Submission C.}\label{res:P3C} 
\editorialdecision{Reject.}
\refereereportlink{03}{3}{C}

The proof attempt rests on two key lemmas (Lemma 1, a Hoeffding-type comparison reducing weighted Bernoulli sums to binomial averages, and Lemma 2, a binomial mean-tail estimate). The cited sources do not contain the stated result for either lemma, and it is not even  explained how the statements would follow from the cited references. For Lemma 1, the reviews note that weakening its strict inequality to a non-strict one would yield a form of Csoka's conjecture, which is an \emph{open} conjecture and likely at least as hard as the MMS conjecture to resolve. The remaining argument, which reduces the cases $p \in [0, 1/3]$
to these two lemmas, is straightforward and appears correct. Overall the submission offers no correct idea beyond this reduction, and the central connection to the Manickam-Miklos-Singhi conjecture is never made. Both reviewers recommended rejection.

\paragraph{Submission D.}\label{res:P3D} 
\editorialdecision{Reject.}
\refereereportlink{03}{3}{D}

The submission proves the inequality for $p=1/k, k \geq 1$ by a coupling argument that directly replicates the one of Alon, Emek, Feldman, and Tennenholtz, without citing them. The submission additionally asserts an incorrect conjecture, that $p=1/k$ and $p=0$ are the only universally good values. Both reviewers recommended rejection.

\subsection{Problem 4}

\paragraph{Submission A.}\label{res:P4A}
\editorialdecision{Reject.}
\refereereportlink{04}{4}{A}

The first half of this solution makes non-trivial progress towards the result, beyond the existing literature, and makes very clear that it does not establish the main result. It can thus be considered as a partial solution. As one reviewer put it, ``[t]his submission contains some of the ideas in the author's solution, but not the most novel idea in the solution.'' At the end, it reduces the main result to a complicated inequality, whose correctness is not argued, and which the reviewers do not view as promising, since the submission ``doesn't give any justification for why this inequality should be true or should be easier to prove.''

\paragraph{Submission B.}\label{res:P4B}
\editorialdecision{Reject.}
\refereereportlink{04}{4}{B}

This is an unusual submission. It proves a different, weaker theorem than the one posed  (one that follows from standard facts) and so does not lead anywhere on the actual problem. It is a verbose solution proving many trivialities: it dwells on an intermediate filling estimate that, as reviewer noted, ``shows up in other submissions and in the problem solution, but this submission doesn't use it to prove anything.'' It does mention possibly interesting ideas at the end, but with no real conclusion.

\paragraph{Submission C.}\label{res:P4C}

\editorialdecision{Reject.}
\refereereportlink{04}{4}{C}

This solution hallucinates that a key result exists in a previous paper by the author of the problem: it states a ``Quoted Theorem'' which is essentially equivalent to the problem, and attributes it to Proposition 9.1 of Guth's ``Area-expanding embeddings of rectangles''~\cite{Guth}, which contains no such statement. As one reviewer put it, the submission ``hallucinates a theorem from the literature which exactly matches the theorem it was asked to prove,'' and then ``trivially deduces the desired theorem from the one it hallucinated.'' The derivation is then tautological. All three reviewers identified this.

\paragraph{Submission D.}\label{res:P4D}

\editorialdecision{Reject.}
\refereereportlink{04}{4}{D}

Within the first page of this solution, two inequalities are stated which are both  false. A scaling argument shows that the two sides scale at different rates, as noted by a reviewer:  ``scaling $R$ and $S$ by $c$ scales the left sides of (1) and (2) by $c^4$ and the right sides by $c^5$.'' There is some evidence in the text that the model has identified an error in its proposed solution; another reviewer granted that ``it does have the merit that it seemed to realize that it went astray.''

\subsection{Problem 5}

\paragraph{Submission A.}\label{res:P5A}
\editorialdecision{Essentially Flawless.}
\refereereportlink{05}{5}{A}

The proof is correct and novel. Its central technical ingredient is the drift estimate of Lemma 1, which applies the stochastic sewing lemma of Khoa Lê to control the drift using carefully chosen time splitting approximations whose terms satisfy the required estimates against the noise filtration. With this estimate in hand, the rest of the argument is a classical ergodic-theory route; the drift is a ``small" perturbation, so a Girsanov entropy bound yields absolute continuity of the transition probabilities with respect to the Gaussian invariant measure of the linear dynamics, from which uniqueness follows. Notably, the submission establishes a somewhat stronger intermediate result than the human solution (finite-time absolute continuity rather than the asymptotic strong Feller property), and it uses the sewing lemma in a different way. All three reviewers rated the submission essentially flawless, but note that the exposition could be improved.  For example, the important role of Lemma 1 should be noted before its proof, some notation is left undefined, and the Lê sewing paper should be cited inline where the lemma is used. 

\paragraph{Submission B.}\label{res:P5B}

\editorialdecision{Reject.}
\refereereportlink{05}{5}{B}

This submission proves that the Gibbs measure is invariant for the SPDE at hand using standard methods; namely, studying the invariant measures of the approximate equations and passing to the limit.  The reviewers note that this result is already known, and should have been cited as such.   

The submission assembled a collection of conditions that would imply uniqueness of the invariant measure, but it did not establish that any of the conditions hold for the equation at hand. The submission acknowledges this gap.  Overall, the submission contains little that is novel.  All three reviewers recommend rejection.

\paragraph{Submission C.}\label{res:P5C}
\editorialdecision{Reject.}
\refereereportlink{05}{5}{C}

Although this submission claims a complete and unconditional proof, it contains a serious error.  

The natural strategy is to first  establish existence of an invariant measure, and then its uniqueness.  Existence follows fairly easily from classical arguments and is handled correctly, although the Gibbs measure used is already known and should be cited as such. The uniqueness argument, however, is flawed. As one reviewer explains, "the AI incorrectly cites a result concerning the resolvent, claiming absolute continuity with respect to an invariant measure everywhere rather than almost everywhere." The submission itself flags pointwise absolute continuity as the "key point", so the incorrect citation sweeps the central difficulty under a rug. This error would be subtle for a non-expert to isolate.  The submission does not introduce substantial new ideas, and all three reviewers recommend rejection.

\paragraph{Submission D.}\label{res:P5D}
\editorialdecision{Reject.}
\refereereportlink{05}{5}{D}

The submission is explicitly a partial solution.  It provides a construction of an invariant measure, but this part of the proof is simple and known in the literature 
(the submission does not provide the citation).  The submission is transparent about leaving large gaps in remaining arguments regarding topological irreducibility and uniqueness of the constructed invariant measure. Overall, this submission does not contain novel results.

\subsection{Problem 6}
\label{sec:P6}
\paragraph{Submission A.}\label{res:P6A} No submission.

    System A did not produce any output on P6. The reason is a series of errors\footnote{e.g. \texttt{Exception: OpenAI response.status=failed with no output items. Error: server\_error An error occurred while processing your request. You can retry your request, or contact us through our help center at help.openai.com if the error persists.}} from the OpenAI API which exceeded the harness's timeout. As System A is based on ChatGPT 5.5 Pro and did at least as well (with regards to referee recommendations) as ChatGPT 5.5 Pro on every other problem, we find it plausible that System A would have received a decision of essentially flawless on P6 without this technical issue. 

\paragraph{Submission B.}\label{res:P6B}
\editorialdecision{Essentially Flawless.}
\refereereportlink{06}{6}{B}

All three reviewers judged the argument mathematically correct and complete, with only expository concerns remaining. One reviewer rated it ``essentially flawless'' on a technical level, but cautioned that ``the fairly technical arguments are sufficiently interleaved and unmotivated that it is quite difficult to check line-by-line.'' Another agreed there were no mathematical issues requiring correction, noting that ``the writing has a tendency to just power through things stream of consciousness style, in a way that is pretty routine to check but is not the
most useful for providing conceptual organization and insight,'' and a third confirmed it ``contains a complete solution in which all the details appear correct.''

\paragraph{Submission C.}\label{res:P6C}
\editorialdecision{Essentially Flawless.}
\refereereportlink{06}{6}{C}

All three reviewers found the solution mathematically correct and complete, with the proof following the same general strategy as the known human solution, but as one reviewer noted ``...with
the conceptual underpinnings stripped out''. One reviewer rated it ``essentially flawless,'' observing that ``all claims made are true, and none cannot be justified with any more than a bit of looking back in the text.'' Another concurred that ``mathematically, it is essentially flawless and does not requi[r]e any factual corrections,'' while suggesting stylistic changes before publication; the remaining reviewer likewise judged that, although ``not well written,'' the submission ``contains a complete solution in which all the details appear correct.''

\paragraph{Submission D.}\label{res:P6D}\editorialdecision{Reject.}
\refereereportlink{06}{6}{D}

After listing some relevant definitions and preliminary calculations, the
submission breaks the proof into cases and solves an easy case, but does not
make progress or introduce nontrivial ideas towards the crux of the problem.  The reviewers did
not find mathematical errors until the end, when an unjustified claim is used
to declare victory.  The writing and notation are confusing but can ultimately
be understood. All three reviewers recommend rejection.

\subsection{Problem 7}

All submissions solved the problem by exhibiting a counterexample and showing that the statement is false. No solution produced a figure illustrating its argument, which a mathematician would expect in this area. There were minor variations in the counterexamples produced. All solutions provided more details about standard facts in the subject than would be required; at that level of detail, some of the less standard arguments were missing.

\paragraph{Submission A.}\label{res:P7A}
\editorialdecision{Essentially Flawless.}
\refereereportlink{07}{7}{A}

Some reviewers noted that this solution was closest to how a mathematician would write a paper. As one reviewer put it, this was ``the closest facsimile to one that might be written by an actual mathematician,'' and another noted that proving the result ``by observing that the CW complex is homeomorphic to $S^2$'' is the approach ``most similar to what most topologists would choose.'' This approach led to a result which is stronger than that of the other submissions. There is an unusual choice of terminology of "graph with poles" that the reviewers had to appropriately interpret. 

\paragraph{Submission B.}\label{res:P7B}
\editorialdecision{Minor Revisions.}
\refereereportlink{07}{7}{B}

This is the only submission where the reviewers reached consensus that the solution would not meet publication standards, despite the fact that the arguments are correct. Stylistically, it is far too long, and its attempts at including citations can be misleading: it cites two standard topology texts (Munkres~\cite{Munkres} and Hatcher~\cite{Hatcher}) for well-known facts, in a way that implies the cellular homology theorem is due to Hatcher.  One reviewer noted that the solution ``introduces the word `admissible' and then uses it extensively, without ever defining what this word means,''  The reviewers interpreted it to refer to the condition introduced by the author in the problem statement, and under this interpretation the solution was deemed to be mathematically correct. Another reviewer judged it ``the one most poorly written --- least like a mathematician.''

\paragraph{Submission C.}\label{res:P7C}
\editorialdecision{Essentially Flawless.}
\refereereportlink{07}{7}{C}

This was viewed by some reviewers as closest to how mathematicians want results to be written. The homology computation used one of many frameworks for computing homology, but a reviewer pointed out that  the choice of framework ``isn't stated anywhere except for a single passing use of the word `cellular' at the end.'' This made the submission difficult to verify. One reviewer ranked it among ``the top two well-written ones --- most like a mathematician.'' 

\paragraph{Submission D.}\label{res:P7D}
\editorialdecision{Minor Revisions.}
\refereereportlink{07}{7}{D}

This solution used notation that would be considered incorrect by most mathematicians. This solution used unusual adverbs in its writing; one reviewer singled out the clause ``it structurally mandates $b_2 \geq 1$'' as ``especially egregious and sounds quite unnatural.'' The proof was structured into parts, which would have been clearer if it had been decomposed into Lemmas.

\subsection{Problem 8}
\paragraph{Submission A.}\label{res:P8A}
\editorialdecision{Reject.}
\refereereportlink{08}{8}{A}

The referees debated about whether this submission should be considered a \textbf{rejection} or would instead require \textbf{major revisions}. 
There is a major gap in the proof, and it is unclear a priori how to fill the gap.  In particular, the assertion that equation (1) describes a flag of valuated matroids is claimed to be a fact which appears in the literature.  The solution includes false citations at this point: one citation proves the fact in a special case, and the other citation is irrelevant.  However, this "fact" is not in the literature.  The referees think that this "fact" is probably true but agreed that proving it requires substantial work.  (They spent a lot of time thinking about this and came up with some ideas for how to do so.)

\paragraph{Submission B.}\label{res:P8B}
\editorialdecision{Major Revisions.}
\refereereportlink{08}{8}{B}

 Both referees felt that this submission needs \textbf{major revisions}. They pointed out major gaps and/or incomprehensible explanations in both Lemmas 5 and 8.   The solution definitely contained ideas that made progress towards the desired goal; with substantial effort, a human expert could potentially augment these ideas and create a complete proof.  There were other lemmas in the submission that had correct but obscure or confusing proofs.  
The referees did not find any serious citation issues with this submission.

\paragraph{Submission C.}\label{res:P8C}
\editorialdecision{Reject.}
\refereereportlink{08}{8}{C}

All  refereees agreed that this solution should be \textbf{rejected}.  The writing style was generally lucid and readable  -- but the most crucial step of the proof was missing.  
The solution spends a long time on easy material and then where it should be providing a substantial amount of justification, it includes just a very vague paragraph that is nowhere near sufficient -- it claims that something is true by standard arguments (which are not sufficient), and includes multiple citations to important books/papers that are important in the field but not relevant to this claim.

\paragraph{Submission D.}\label{res:P8D}
\editorialdecision{Reject.}
\refereereportlink{08}{8}{D}

All referees agreed that this submission would be \textbf{rejected} right away, in part because it attempted to prove a false statement.  Moreover, the solution meandered and contained sentences that the referees simply could not parse.  Insofar as anything could be extracted, the solution performed some calculations that are not useful.

\subsection{Problem 9}
\paragraph{Submission A.}\label{res:P9A}
\editorialdecision{Minor Revisions.}
\refereereportlink{09}{9}{A}

Both referees recommended  minor revisions.  They felt that the combinatorial interpretation provided (in terms of ordered set partitions) was the nicest among the four submissions. 
This combinatorial interpretation was felt to be simpler -- but a bit more contrived and less natural -- than the one given in the human solution.
{\it This solution was different from the human solution and other AI solutions}

The solution provides the correct references  in the bibliography, but the references are not cited in the text; instead, the text claims that some results are "standard" when in fact they appear in recent papers.  The fact that citations are not given in the text makes it more difficult to check the solution  for correctness.  

The presentation is quite dense: all the steps are included in the same proof.  The solution would be more readable if the proof were subdivided into smaller steps/lemmas.
The proof used some nonstandard notation, i.e. $H_n$ for the Hilbert series.

\paragraph{Submission B.}\label{res:P9B}
\editorialdecision{Minor Revisions.}
\refereereportlink{09}{9}{B}

All three referees recommended
 minor revisions.
The combinatorial interpretation provided 
(which involves a permutation
and a partition  that are not genuinely related) seems slightly unnatural.
{\it This solution is different from the human solution and the other AI solutions.}

Parts of the proof were roundabout and hence difficult to follow; however, the solution didn't contain technical errors.  The citations were mostly appropriate but in same cases the exposition could have been shortened by citing classical results.  The solution didn't seem to be aware of a Rhodes-Wilson paper which would have simplified the proof.  
The use of notation was inconsistent -- there was one quantity that was sometimes called "maj" and sometimes called "m";
similarly with "des" and "d".
The proof also included unnecessary and trivial details, 
such as "because $(d-1)-b=d-b-1.$"

\paragraph{Submission C.}\label{res:P9C}
\editorialdecision{Minor Revisions.}
\refereereportlink{09}{9}{C}

Both referees recommended 
 minor revisions.
The proposed solution gives a combinatorial formula for the coefficients in terms of triples
$(r, \pi, t)$, where $r$ and $t$ are integers and $\pi$ is an ordered set partition; having the $t$ there makes the combinatorial interpretation slightly unnatural. 
The solution is different from the human solution and the other AI solutions.

The formula is correct and the solution is mostly well written except for some citation issues.  The proof of Lemma 3 clearly uses work of Lentfer but doesn't cite it; other parts of the proof cite relevant sources but the citations appear several sentences before or after they should.  Another citation 
refers to a theorem that exists but the citation doesn't properly locate that theorem in the paper where it appears.  Lemma 2 is stated and proved without citation even though that statement and its proof appear elsewhere in the literature.
The proof used some nonstandard notation, i.e. $H_n$ for the Hilbert series.

\paragraph{Submission D.}\label{res:P9D}
\editorialdecision{Reject.}
\refereereportlink{09}{9}{D}

All three referees agreed that 
this submission should be rejected.
The submission initially declares that it only has a partial solution, but much of the rest of the writeup is written as if it has a complete solution.  The submission does not give a combinatorial formula, but only suggests a strategy that might lead
to one. However, the steps of this strategy are not explained clearly, do not seem to be well
connected, and references are missing.  There are some nonsensical statements about e.g. evaluating a "ring at a vector space"
(Step 4), or "evaluating a $GL_n$ character at a superspace".
The final remark on the case $d = 0$ is correct, but no reference is given.   Some uses of the words/phrases "exactly", "strictly", and "mathematically identical" appeared in unnatural places.
There is no bibliography, and only  one in-line citation, and this citation is not to the correct paper.

\subsection{Problem 10}
\paragraph{Submission A.}\label{res:P10A}
\editorialdecision{Major Revisions.}
\refereereportlink{10}{10}{A}

The two reviewers who recorded a recommendation both called for major revisions, agreeing that the overall approach is correct but the execution falls well short of publishable form. One reviewer noted that ``while the approach is correct, it would take an expert in the subject matter anywhere from 2 days to perhaps a week of work to rewrite this to fix all the issues,'' adding that as written ``it is not really understandable except to the limited number of experts'' in the area. The other identified concrete gaps---``the usage of operator bimodules in the definition of the bidual boundary piece is missing, and additionally the bimodule computations are very confusing and unjustified''---but allowed that ``it is possible that there may be a path towards publication for this article.''

\paragraph{Submission B.}\label{res:P10B}
\editorialdecision{Reject.}
\refereereportlink{10}{10}{B}

Both reviewers who assessed this submission recommended rejection, on the grounds that it does not address the actual problem and lacks basic rigor. One reviewer wrote that the writeup ``fails to meet even the most basic standards of rigor, and additionally does not answer the problem at hand,'' and the other recommended rejection ``as it fails to solve the problem.''

\paragraph{Submission C.}\label{res:P10C}
\editorialdecision{Major Revisions.}
\refereereportlink{10}{10}{C}

All three reviewers recommended major revisions: the high-level strategy is coherent, but the central technical lemmas are not actually established. One reviewer placed the submission ``between requiring major revisions and outright rejection,'' observing that it ``does not prove at all the main technical result needed, instead citing a paper and supplying no hints as to how to apply'' it, and that while ``this approach works, this is by no means obvious or routine and requires a significant amount of additional arguments.'' Another noted that correct proofs of Lemmas~3.1 and~3.2 are required, but that ``the rest of the strategy seems to be coherent, which is why I recommend a major revision, as opposed to a straightforward rejection.''

\paragraph{Submission D.}\label{res:P10D}
\editorialdecision{Reject.}
\refereereportlink{10}{10}{D}

The reviewers recommended rejection because the submission rests on a fabricated definition. As one reviewer put it, the solution ``hallucinates a definition of proper proximality that is plainly wrong and justifies this by citing a paper from before the concept was even defined for general tracial von Neumann algebras,'' so that ``the strategy pursued is thus unrelated to the problem altogether.'' The same reviewer observed that ``the AI model seems to have troubles determining and accessing the correct references,'' and another simply recommended rejection ``as the proof is wrong.''

\section{Supplementary documents}

We are hosting supplementary documents on the 
\href{https://github.com/1stproof/batch-2/}{First Proof GitHub page}.  
\begin{itemize}
    \item \href{https://github.com/1stproof/batch-2/tree/main/batch-2-human-solution}{The human-generated solutions to the problems}
    \item \href{https://github.com/1stproof/batch-2/tree/main/batch-2-AI-solutions}{The AI-generated solutions to the problems}
    \item \href{https://github.com/1stproof/batch-2/tree/main/batch-2-reviews}{The referee reports on the AI-generated solutions}
        \item \href{https://github.com/1stproof/batch-2/tree/main/batch-2-raw-outputs}{The logs from the AI systems}
        \item \href{https://github.com/1stproof/batch-2/tree/main/batch-2-submissions}{Source code for the harnesses created by the academic teams}
\end{itemize}

\bibliographystyle{plain}
\bibliography{references}

\clearpage
\renewcommand{\notesname}{Email addresses}
\setlength{\parindent}{0pt}
\def\enotesize{\small}
\theendnotes

\end{document}